\pdfoutput=1
\documentclass[10pt,journal]{IEEEtran}
\IEEEoverridecommandlockouts
\usepackage{microtype}
\usepackage{graphicx}
\usepackage{xcolor}
\usepackage{subfigure}
\usepackage{booktabs}
\usepackage{hyperref}
\usepackage{array}
\usepackage{multirow}
\usepackage{algorithm}
\usepackage{algorithmic}
\usepackage{amsfonts}
\usepackage{amsmath}

\newcommand{\red}[1]{\textcolor[rgb]{1.0, 0.1, 0.5}{\textbf{#1}}}

\newcommand{\todo}[1]{#1}

\newcommand{\LWC}{LWC}
\newcommand{\FLWC}{F-LWC}

\def\BibTeX{{\rm B\kern-.05em{\sc i\kern-.025em b}\kern-.08em
    T\kern-.1667em\lower.7ex\hbox{E}\kern-.125emX}}
    
\begin{document}
\title{Compute-in-Memory based Neural Network Accelerators for Safety-Critical Systems: Worst-Case Scenarios and Protections}

\author{Zheyu~Yan,~\IEEEmembership{Member,~IEEE,}
        Xiaobo~Sharon~Hu,~\IEEEmembership{Fellow,~IEEE,}
        and~Yiyu~Shi,~\IEEEmembership{Senior~Member,~IEEE}
\thanks{Z. Yan, X. S. Hu, and Y. Shi are with the Department
of Computer Science and Engineering, University of Notre Dame, Notre Dame, IN, 46556 USA 
(e-mail: \{zyan2, shu, yshi4\}@nd.edu).}
\thanks{Please address comments to zyan2@nd.edu and yshi4@nd.edu.}
}

\markboth{IEEE Transactions on Computer Aided Design of Integrated Circuits \& Systems,~Vol.~1, No.~1, April~2023}%
{Jia \MakeLowercase{\textit{et al.}}: A Sample Article Using IEEEtran.cls for IEEE Journals}

\maketitle

\begin{abstract} 
Emerging non-volatile memory (NVM)-based Computing-in-Memory (CiM) architectures show substantial promise in accelerating deep neural networks (DNNs)  due to their exceptional energy efficiency. However, NVM devices are prone to
device variations.
Consequently, the actual DNN weights mapped to NVM devices can differ considerably from their targeted values, inducing significant performance degradation. Many existing solutions aim to optimize average performance amidst device variations, which is a suitable strategy for general-purpose conditions. However, the worst-case performance that is crucial for safety-critical applications is largely overlooked in current research. In this study, we define the problem of pinpointing the worst-case performance of CiM DNN accelerators affected by device variations. Additionally, we introduce a strategy to identify a specific pattern of the device value deviations in the complex, high-dimensional value deviation space, responsible for this worst-case outcome. Our findings reveal that even subtle device variations can precipitate a dramatic decline in DNN accuracy, posing risks for CiM-based platforms in supporting safety-critical applications. Notably, we observe that prevailing techniques to bolster average DNN performance in CiM accelerators fall short in enhancing worst-case scenarios. \todo{In light of this issue, we propose a novel worst-case-aware training technique named A-TRICE that efficiently combines adversarial training and noise-injection training with right-censored Gaussian noise to improve the DNN accuracy in the worst-case scenarios. Our experimental results demonstrate that A-TRICE improves the worst-case accuracy under device variations by up to 33\%.}
\end{abstract}

\section{Introductions}

Deep Neural Networks (DNNs) have achieved unparalleled success in various perception tasks, such as object detection, speech recognition, and image classification. As a result, there is a growing trend to harness DNNs for edge applications, which span smart sensors, smartphones, automobiles, and the like~\cite{yang2020co, sheng2022larger}. However, the limited computational resources and power constraints of edge platforms mean that CPUs or GPUs may not always be the optimal choice for deploying computation-intensive DNNs on these devices.

An appealing alternative for edge DNN deployment is the Compute-in-Memory (CiM) DNN accelerators~\cite{shafiee2016isaac}. They offer the advantage of minimizing data movement through in-situ weight data access~\cite{sze2017efficient}. By incorporating non-volatile memory (NVM) devices, such as phase-change memories (PCMs), resistive random-access memories (RRAMs), and ferroelectric field-effect transistors (FeFETs), CiM can outperform conventional MOSFET-based designs in terms of memory density and energy efficiency~\cite{chen2016eyeriss}. Nonetheless, the reliability of NVM devices remains a concern. Specifically, there exist issues including device-to-device variations stemming from fabrication defects, and cycle-to-cycle variations due to devices' inherent unpredictability. This stochastic behavior causes device value deviations from the targeted device value (\emph{e.g.}, targeted device conductance used to represent DNN weights) and finally results in the difference of weight values in DNN inference. If not addressed aptly, the deviation of weight values during computations from their targeted values can lead to a pronounced decline in performance.

To assess the robustness of CiM DNN accelerators, researchers often resort to a Monte Carlo (MC) simulation-based evaluation approach~\cite{peng2019dnn+}. A device variation model is derived from physical measurements. In every MC iteration, a device instance is sampled from the variation model, and the DNN performance metrics are recorded. This process is repeated multiple times, often thousands until the DNN performance distribution stabilizes. Current methodologies~\cite{yan2020single, jin2020improving, liu2019fault, he2019noise, yan2022swim} typically involve up to 10,000 MC iterations, making the process highly time-intensive. Some researchers employ Bayesian Neural Networks (BNNs) for robustness evaluation against device variations~\cite{gao2021bayesian}. However, the variational inference process in BNNs is essentially a type of MC simulation.

Numerous strategies have been introduced to enhance the average performance of CiM DNN accelerators under the influence of device variations. These strategies generally bifurcate into two main categories: (1) minimizing device value deviations, and (2) improving DNN robustness. A widely-used method to reduce device value deviation is write-verify~\cite{shim2020two}, where iterative write and read (verify) operations ensure that the final difference between the weights programmed into the devices and the intended values remains within a pre-defined limit. Employing write-verify has proven effective, reducing weight value differences to below 3\% and limiting average DNN performance reduction to less than 0.5\%~\cite{shim2020two}. To enhance DNN robustness, there is a plethora of techniques. Neural architecture search~\cite{yan2021uncertainty, yan2022radars}, for instance, has been designed to automatically sift through a specified search space to identify more resilient DNN architectures. Variation-aware training~\cite{jiang2020device, he2019noise} introduces device variation-induced weight fluctuations during training, ensuring the resulting DNN weights are fortified against similar perturbations. Additional strategies encompass on-chip in-situ training~\cite{yao2020fully}, which directly trains DNNs on inherently noisy devices, and Bayesian Neural Network (BNN) methods that leverage the BNN's variational training process to enhance DNN robustness~\cite{gao2021bayesian}.

\begin{figure}[ht]
    \includegraphics[trim=0 150 130 0, clip, width=0.95\linewidth]{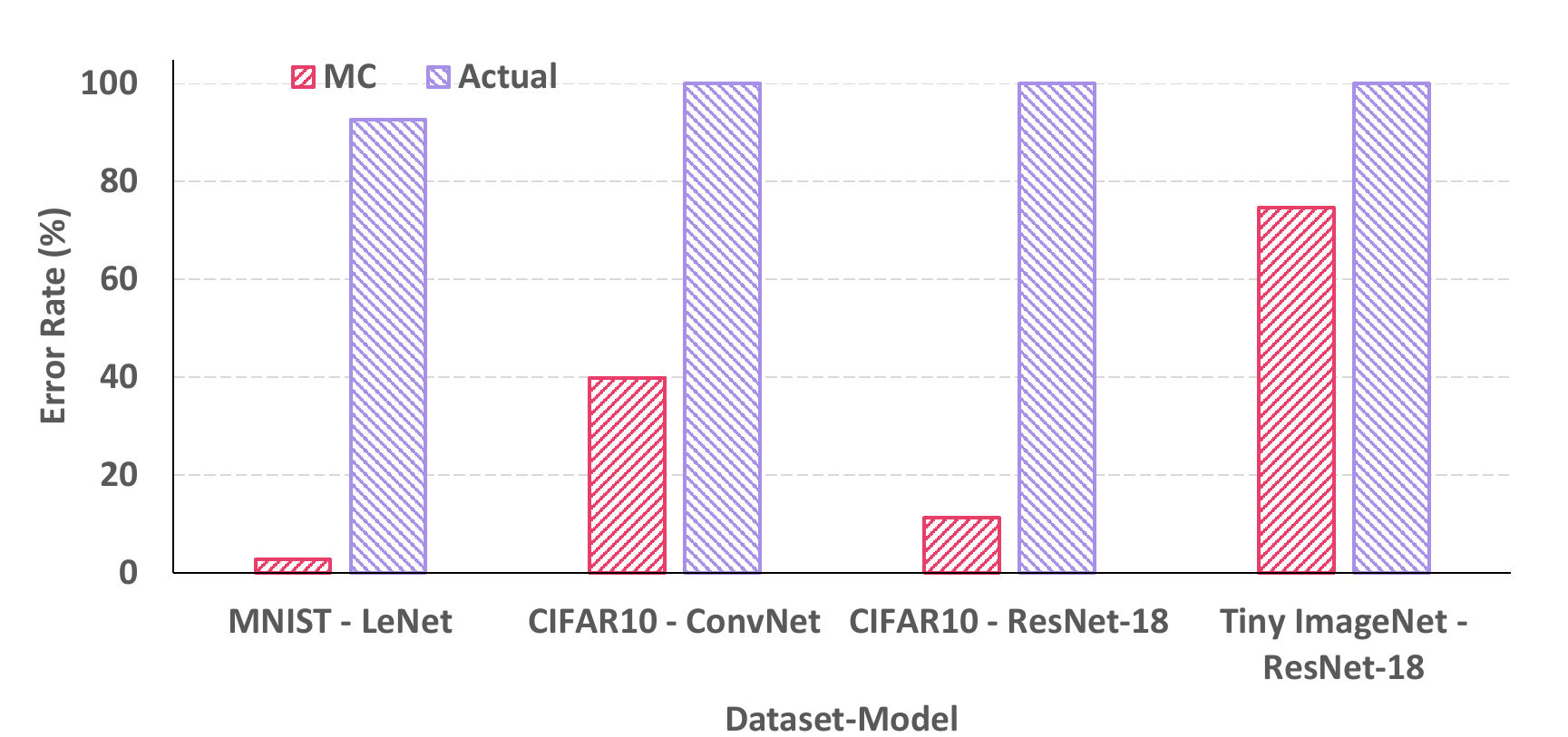}
    \caption{
    Comparison of the highest top-1 error (\emph{i.e., worst-case}) identified through 100K MC simulations versus our method, using the device variation model detailed in Section~\ref{sect:model}. Results are presented for four DNNs: LeNet (MNIST), CovNet (CIFAR-10), ResNet-18 (CIFAR-10), and ResNet-18 (Tiny ImageNet).
    }
    \label{fig:Intro}
\end{figure}

All the evaluation and robustness enhancement methods commonly target the average performance of a DNN with the influence of device variations, suitable for general-purpose conditions. However, in safety-critical applications where failure might lead to loss of life (such as in nuclear systems, aircraft flight control, and medical devices), significant environmental harm, or property damage, relying solely on average performance is insufficient. It is essential to consider the worst-case performance, irrespective of its probability~\cite{wang2022efficient}. Addressing this concern presents a challenge: Due to the extremely high dimensionality of the device value deviation space, hoping to capture the worst-case scenario via MC simulations is infeasible. As depicted in Fig.~\ref{fig:Intro}, for various DNNs across different datasets, even after convergence is achieved with 100K MC runs, the highest discovered DNN top-1 error rate remains notably higher compared to what our method (to be discussed later in this paper) identifies, where the worst-case error approaches 100\%.

Despite the importance of the problem, it remains largely overlooked in existing research. The scarce related work approaches the problem from a security perspective~\cite{tsai2021formalizing}, where a weight-projected gradient descent (PGD) method identifies weight perturbations that cause input misclassifications. However, this strategy aims to generate a successful weight perturbation attack rather than pinpoint the worst-case scenario amidst all potential variations.

To fill the gap, we introduce an optimization-based analysis framework in this paper, tailored to efficiently identify the worst-case performance of DNNs in a CiM accelerator, in the scenario where the maximum weight variations are bound by write-verify. Our research highlights that this challenge can be delineated as a constrained optimization problem with a non-differentiable objective. Remarkably, it is amenable to relaxation and the finally relaxed problem can be resolved through gradient-based optimization. \todo{Because this optimization problem is solved by a Lagrange multiplier-based method, we name our framework of evaluating the worst-case performance to be Lagrange-based Worst-Case analysis (\LWC).} Our experiments span various networks and datasets, especially focusing on a prevalent setup where each device encodes 2 bits of data under write-verify, producing a peak weight perturbation magnitude of 0.03~\cite{shim2020two}. Notably, Weight PGD~\cite{tsai2021formalizing}, the only method of relevance in the literature, can unearth worst-case scenarios where the accuracy mirrors that of a random guess. In contrast, our approach can unveil scenarios where accuracy verges on zero. \todo{
However, our method (\LWC) requires iterative tuning of the Lagrange multiplier, making it inefficient to use if it needs to be called upon multiple times (\emph{e.g.}, in hyper-parameter tuning). Therefore, we further propose a rapid worst-case evaluation scheme named \FLWC, based on the fast gradient sign method (FGSM). \FLWC~can efficiently estimate the worst-case performance of DNN models with the influence of device variations, offering an evaluation precision comparable to \LWC~with a 2x speedup and further offering a 10x speedup with a reasonable evaluation precision.
}

\todo{
Moreover, as highlighted in our previous work~\cite{yan2022computing}, the existing methods aiming at improving the robustness of DNNs under the influence of device variations do not effectively enhance the worst-case performance. Notably, the two primary solutions, reducing device value deviations through more rigorous write-verify procedures and implementing variation-aware training—either incur significant overheads or fall short in terms of effectiveness, underscoring the need for additional research. Our recent study~\cite{yan2023improving} demonstrates that a unique noise-injection training approach, which introduces right-censored Gaussian (RCG) noise to DNN weights during training, can boost the quantile (\emph{e.g.}, median) performance of DNN models under the influence of device variations. Based on this discovery, in this paper, we introduce a novel worst-case-aware training strategy named \underline{A}dversarial \underline{T}raining with \underline{RI}ght-\underline{C}ensored Gaussian Nois\underline{E} (A-TRICE). This strategy effectively improves the worst-case performance of DNN models. We first present a novel adversarial training approach based on our rapid worst-case evaluation method~\FLWC. Subsequently, we merge the principles of RCG noise injection training and adversarial training to establish our comprehensive A-TRICE framework.
}

The major contributions of our study can be encapsulated as:
\begin{itemize}
    \item We are the first that formulate the problem of finding the worst-case performance in DNN CiM accelerators amidst device variations.
    \item We introduce an adept gradient-based methodology \LWC~to navigate the intricacies of the non-differentiable optimization problem and determine the worst-case scenario. 
    \item Our experiments underscore the uniqueness of our framework in identifying the DNN's worst-case performance.
    \item We emphasize that, even when maximum weight perturbations are stringently constrained (as with write-verify), DNNs can still undergo a substantial dip in accuracy. Consequently, CiM accelerator deployments in safety-critical environments demand vigilance.
    \item \todo{We propose a novel technique named \FLWC~based on the fast-gradient sign method to efficiently estimate the worst-case performance. \FLWC~achieves similar worst-case performance evaluation with a 2x speedup.}
    \item \todo{We develop a novel training method A-TRICE that improves the worst-case performance by up to 33\%.}
\end{itemize}

The structure of this paper unfolds as follows: Section~\ref{sect:related} delves into the background information of CiM DNN accelerators, the robustness challenges spurred by device variations, and the current solutions addressing these concerns. Subsequent to this, in Section~\ref{sect:proposed}, we formulate the problem of detecting the DNN's worst-case performance amidst device variations and propose a framework \LWC~for its resolution, corroborated by experimental evidence. In Section~\ref{sect:protect}, we first propose a fast evaluation framework \FLWC~to estimate the worst-case performance so that we can then evaluate the effectiveness of existing methods in improving the worst-case performance. In light of their in-effectiveness, we further discuss our proposed method A-TRICE, and its effectiveness in the same section. We culminate our discourse with conclusive observations in Section~\ref{sect:conclusion}.
\section{Related Works}\label{sect:related}
In this section, we first describe the overall structure and major building blocks of CiM DNN accelerators, then discuss their robustness challenges spurred by device variations, and finally introduce the current solutions addressing these concerns. 

\subsection{Crossbar-based CiM Engine}\label{sec:2.1}
\begin{figure}[ht]
\begin{center}
\centerline{\includegraphics[trim=0 150 550 0, clip, width=0.4\linewidth] {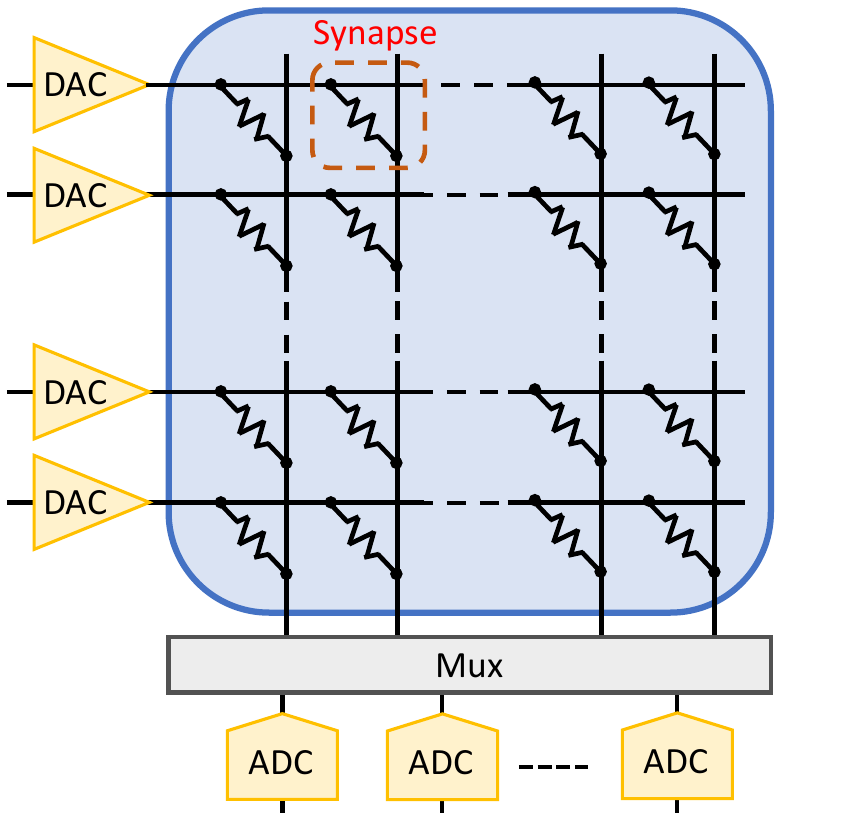}}
\caption{
Depiction of the crossbar architecture: Inputs are introduced horizontally and interact with weights preserved in the NVM devices at each intersection. These multiplicative interactions are aggregated vertically, with the cumulative result functioning as the output.
}
\label{fig:crossbar}
\end{center}
\end{figure}

The crossbar array serves as the major computational component of CiM DNN accelerators. This array can execute matrix-vector multiplications within a single clock cycle. In such an array, matrix values (\emph{e.g.}, DNN weights) are stored at the intersection of vertical and horizontal lines using NVM devices like RRAMs and FeFETs. Vector values, akin to DNN inputs, are introduced through horizontal data lines as voltage. The resultant output is then channeled through vertical lines in a current form. Computation within the crossbar array operates in the analog domain, following Kirchhoff's laws. However, for other crucial DNN operations such as pooling and non-linear activation, peripheral digital circuits are employed. Thus, digital-to-analog and analog-to-digital converters (DACs and ADCs) bridge these different components, specifically with DACs transforming digital input data into voltage and ADCs converting analog output currents back to digital values.

Resistive crossbar arrays are susceptible to a variety of variations and noises. The predominant sources include spatial and temporal variations. Spatial variations, stemming from fabrication imperfections, exhibit both localized and widespread correlations across devices. Additionally, due to the unpredictability inherent in device materials, NVM devices experience temporal variations, leading to conductance fluctuations when programmed across different instances. Typically, these temporal variations are distinct between devices and are not influenced by the intended programming value~\cite{feinberg2018making}. For the scope of this study, we treat the effects of these non-idealities as independent across each NVM device. Nonetheless, our proposed framework can be adapted to address other variation sources with some alterations.

\subsection{Evaluating DNN Robustness Against Device Variations}

The prevailing approach in current research utilizes Monte Carlo (MC) simulations to evaluate the resilience of CiM DNN accelerators against device variations. Initially, models for device variation and circuitry are established based on physical measurements. The DNN being assessed is then mapped onto the circuit model, facilitating the determination of the desired value for each NVM device. In each MC iteration, for every device, a sample of its non-ideal state is drawn from the variation model to set the actual conductance value. Subsequently, metrics reflecting DNN performance, like classification accuracy, are recorded. This process is repeated, often thousands of times, until a consistent DNN performance distribution is achieved. Conventional methods cited in existing studies~\cite{yan2020single, yan2021uncertainty, he2019noise, liu2019fault, yan2022swim} typically involve around 10,000 MC iterations, an extremely time-consuming process. Empirical studies~\cite{yan2020single, yan2021uncertainty} suggest that 10k MC cycles are adequate for determining average DNN accuracy, though no concrete theoretical guarantees are provided.

Other scholars have also probed the effects of weight perturbations on the security of neural networks~\cite{wu2020adversarial, tsai2021formalizing}. Termed ``Adversarial Weight Perturbation'', this research avenue seeks to relate weight perturbations to the well-researched topic of adversarial examples-based attacks, wherein DNN inputs are deliberately altered to induce misclassifications. One study~\cite{wu2020adversarial} trained DNNs using adversarial examples to gather data on adversarial weight perturbation. More recently, \cite{tsai2021formalizing} attempted to identify adversarial weight perturbation using an adapted weight Projected Gradient Descent (PGD) attack. This approach can pinpoint a minor perturbation capable of diminishing DNN accuracy. Nonetheless, while the study concentrates on the attack's success, it does not provide a guaranteed assessment of the worst-case weight perturbation, as will be evidenced by our forthcoming experimental results.

\subsection{Addressing the Impact of Device Variations}\label{sec:2.3}
Several strategies have been introduced to tackle the challenge of device variations in CiM DNN accelerators. This section briefly discusses the two predominant approaches: bolstering DNN robustness and curtailing device variations.

A prevalent technique to bolster DNN robustness in the face of device variations is variation-aware training\cite{jiang2020device,peng2019dnn+,he2019noise,chang2019nv}. This method, sometimes referred to as noise-injection training, integrates variations into DNN weights during the training phase. This integration fosters a DNN model that exhibits statistical resilience to the effects of device variations. In each training iteration, a variation instance, drawn from a predefined distribution, augments the weights during the forward pass, leaving the backpropagation phase unaffected by noise. Upon gradient collection, this variation is discarded, and the pristine weight is updated based on the accrued gradients. Alternatives encompass the design of sturdier DNN architectures~\cite{jiang2020device, yan2021uncertainty, gao2021bayesian} and pruning techniques~\cite{jin2020improving, chen2021pruning}.

In efforts to decrease device variations, the write-verify procedure\cite{shim2020two, yao2020fully} is commonly employed during programming. Initially, an NVM device is programmed to a set state via a designated pulse pattern. Subsequent to this, the device's value is verified to ascertain if its conductance aligns within a stipulated range of the desired value, essentially assessing its accuracy. If discrepancies arise, a supplemental update pulse is initiated to draw the device conductance nearer to the target. This loop persists until the disparity between the programmed device value and the target diminishes to a satisfactory margin, typically taking a handful of cycles. Cutting-edge research suggests that by selectively applying write-verify to a subset of pivotal devices, one can uphold the average accuracy of a DNN \cite{yan2022swim}. Additionally, a variety of circuit design initiatives~\cite{shin2021fault, jeong2022variation} have been put forth to counteract device variations.

\section{Evaluating Worst-Case Performance of CiM DNN Accelerators}\label{sect:proposed}
A primary consequence of device variations is the deviation in the conductance of the NVM devices from their intended values. This discrepancy arises from device-to-device and cycle-to-cycle variations during programming. As a result, the weight values of a DNN experience perturbations, which subsequently influence its accuracy. In Section~\ref{sect:model}, we start by modeling the effects of NVM device variations on weight perturbation, operating under the assumption that write-verify is employed to limit these variations. Building on this weight perturbation model, Section~\ref{sect:problem} formulates the challenge of pinpointing the minimum DNN accuracy in the face of weight perturbations, and we introduce a framework named  Lagrange-based Worst-Case evaluation (\LWC) for its resolution. Section~\ref{sect:method} showcases the experimental findings.

\subsection{Modeling of Weight Perturbation Due to Device Variations}\label{sect:model}

In this section, we present our model that delineates the effects of device variations on DNN weights. Specifically, our primary focus in this paper is on the discrepancies introduced during the programming process, where the conductance value programmed into the NVM devices deviates from the intended value.

For a weight depicted by $M$ bits, its desired value, denoted by $\mathcal{W}_{des}$, is given as
\begin{equation}
    \mathcal{W}_{des} = \sum_{j=0}^{M-1}{m_j \times 2^j}
\end{equation}
where $m_j$ is the value of the $j^{th}$ bit of the intended weight. Given that each NVM device can store data equivalent to $K$ bits, a single weight value in the DNN requires $M/K$ devices for representation\footnote{For simplicity, we assume that M is divisible by K.}. This mapping procedure can be articulated as:
\begin{equation}
    g_i = \sum_{j=0}^{K -1} m_{i\times K + j} \times 2^j
\end{equation}
Here, $g_i$ denotes the intended conductance of the $i^{th}$ device encoding a particular weight. It is pertinent to note that the mapping strategy remains consistent even for negative weights.

Assuming that all devices employ the write-verify technique, the deviation between the genuine conductance of each device and its targeted value remains within specified bounds~\cite{feinberg2018making}:
\begin{equation}
    gp_i = g_i + n_i, s.t. -th\leq n_i \leq th
\end{equation}
In this equation, $gp_i$ represents the actual programmed conductance, and $th$ indicates the permissible threshold set by write-verify.



When programming a weight, the true value, denoted by $\mathcal{W}_{p}$, as stored on the devices, can be described as:

\begin{align}
\begin{split}
    \mathcal{W}_{p}     & = \sum_{i=0}^{M/K -1}2^{i\times K}{gp_i } \\
                        & = \sum_{i=0}^{M/K -1}2^{i\times K}{g_i + n_i} \\
                        & = \mathcal{W}_{des} + \sum_{i=0}^{M/K-1}{n_i \times 2^{i\times K}}\\
    \mathcal{W}_{des} - th_g &\leq \mathcal{W}_{p} \leq \mathcal{W}_{des} + th_g\label{eq:noise}
\end{split}
\end{align}
Here, $th_g = \sum_{i=0}^{M/K-1}{\left(th \times 2^{i\times K}\right)}$ is referred to as the {\em weight perturbation bound} in this paper.

\subsection{Problem Definition}\label{sect:problem}




Having established our weight perturbation model, we can proceed to delineate the task of determining the worst-case accuracy of a DNN. For simplicity, we denote a neural network by $\{f, \mathbf{W}\}$, where $f$ refers to its architecture and $\mathbf{W}$ its set of weights. The forward pass, or computation, of this neural network is symbolized by $f(\mathbf{W},\mathbf{x})$, with $\mathbf{x}$ as its inputs.

According to the model detailed in Section~\ref{sect:model}, weight perturbations due to device variations are additive and independent. This allows us to represent the forward pass of a neural network under the influence of device variation as $f(\mathbf{W} + \Delta\mathbf{W},\mathbf{x})$. Here, $\Delta\mathbf{W}$ symbolizes the weight perturbation induced by these variations. The \textit{perturbed neural network} is then defined as $\{f, \mathbf{W} + \Delta\mathbf{W}\}$.

Based on the above elaborations, we formulate our problem as follows: For a given neural network $\{f, \mathbf{W}\}$ and a reference dataset $D$, identify the perturbation $\Delta\mathbf{W}$ such that the accuracy of the perturbed neural network $\{f, \mathbf{W} + \Delta\mathbf{W}\}$ on dataset $D$ is the lowest, staying within the bounds of permissible weight perturbation. Throughout this document, we will reference this particular perturbation as the \textit{worst-case weight perturbation}. Similarly, the resulting performance (or accuracy) will be termed the \textit{worst-case performance (or accuracy)}, and the neural network associated with it will be the \textit{worst-case neural network}.

Given the problem definition, we can represent it as the following optimization problem:

\begin{equation}
\begin{aligned}
    \underset{\Delta\mathbf{W}}{\mathrm{minimize}} \ & \ |\{f(\mathbf{W}+\Delta\mathbf{W},\mathbf{x}) == t \ | \ (x,t)\in D\}| \\
    \mathrm{subject \ to} \ & \ \mathcal{L}(\Delta\mathbf{W}) \leq th_g\label{eq:ori_def}
\end{aligned}
\end{equation}

Here, $x$ and $t$ symbolize the input data and its corresponding classification label from the dataset $D$, respectively. The term $\mathcal{L}(\Delta\mathbf{W})$ represents the maximum magnitude of the weight perturbation, denoted as $\max(\lvert\Delta\mathbf{W}\rvert)$. The value of $th_g$ is derived from the weight perturbation boundary as discussed in equation (\ref{eq:noise}) from Section~\ref{sect:model}. The notation $|A|$ signifies the cardinality (or size) of set $A$. Given that $f$, $\mathbf{W}$, and $D$ are constants in this scenario, our primary objective is to pinpoint the $\Delta\mathbf{W}$ that diminishes the set of correct classifications to its lowest possible size, thereby leading to the most unfavorable accuracy.

\subsection{Finding the Worst-Case Performance}\label{sect:method}


The optimization problem defined in equation (\ref{eq:ori_def}) is challenging to solve directly due to its non-differentiable objective. In this section, we introduce a framework that reformulates the problem, allowing us to tackle it using existing optimization techniques.

Initially, we consider a slight relaxation of the objective. Assuming that there exists a function $p$ such that the condition $f(\mathbf{W}+\Delta\mathbf{W},\mathbf{x}) == t$ holds true if and only if $p(\mathbf{x}, \{f, \mathbf{W}+\Delta\mathbf{W}\}) \geq 0$~\cite{carlini2017towards}. Consequently, the optimization objective in equation (\ref{eq:ori_def}) 
can be recast as:
\begin{equation}
    \underset{\Delta\mathbf{W}}{\mathrm{minimize}} \sum_{\mathbf{x}\in D} p(\mathbf{x}, \{f, \mathbf{W}+\Delta\mathbf{W}\}). \label{eq:relaxed}
\end{equation}

Intuitively, minimizing the value in equation (\ref{eq:relaxed}) can drive the minimization of the objective in equation (\ref{eq:ori_def}). These two optimization tasks converge in their aims when, under the influence of $\Delta\mathbf{W}$, all data points in $D$ are misclassified.

Several options for \( p(\mathbf{x}, \{f, \mathbf{W}+\Delta\mathbf{W}\}) \) satisfy the given criteria. We present some notable examples below.

{\scriptsize
\begin{equation}
\begin{split}
    \mathbf{O} = & \ f(\mathbf{W}+\Delta\mathbf{W},\mathbf{x}) \\
    \mathbf{Z} = & \ \mathrm{Softmax}(f(\mathbf{W}+\Delta\mathbf{W},\mathbf{x})) \\
    p_1(\mathbf{x}, \{f, \mathbf{W}+\Delta\mathbf{W}\}) = & \ loss(\mathbf{O}, t) - 1\\
    p_2(\mathbf{x}, \{f, \mathbf{W}+\Delta\mathbf{W}\}) = & \max\{O_{t} - \max_{i\neq t}(O_i), 0\}\\
    p_3(\mathbf{x}, \{f, \mathbf{W}+\Delta\mathbf{W}\}) = & \ \mathrm{softplus}(O_t-\max_{i\neq t}(O_i)) - \log(2)\\
    p_4(\mathbf{x}, \{f, \mathbf{W}+\Delta\mathbf{W}\}) = & \max\{O_t-0.5,0\}\\
    p_5(\mathbf{x}, \{f, \mathbf{W}+\Delta\mathbf{W}\}) = & \log(2 \cdot O_t -2)\\
    p_6(\mathbf{x}, \{f, \mathbf{W}+\Delta\mathbf{W}\}) = & \max\{Z_{t} - \max_{i\neq t}(Z_i), 0\}\\
    p_7(\mathbf{x}, \{f, \mathbf{W}+\Delta\mathbf{W}\}) = & \ \mathrm{softplus}(Z_t-\max_{i\neq t}(Z_i)) - \log(2)\\
\end{split}
\end{equation}}
The function \( \mathrm{softplus}(x) \) is defined as \( \log(1+\exp(x)) \), and \( loss(\mathbf{O}, t) \) denotes the cross-entropy loss.

Based on empirical findings, in this paper, we opt for
\begin{equation}
\begin{split}
    p(\mathbf{x}, \{f, \mathbf{W}+\Delta\mathbf{W}\}) = & \max\{\max_{i\neq t}(O_i) - O_{t}, 0\}
\end{split}\label{eq:rel_ch}
\end{equation}
The optimization problem is then relaxed to the form of:
\begin{equation}
\begin{split}
    \underset{\Delta\mathbf{W}}{\mathrm{minimize}} & \ \ \ \  \sum_{\mathbf{x}\in D} p(\mathbf{x}, \{f, \mathbf{W}+\Delta\mathbf{W}\}) \\
    \mathrm{s.t.}     & \ \ \ \  \mathcal{L}(\Delta\mathbf{W}) \leq th_g
\end{split}\label{eq:rel_problem}
\end{equation}

For the relaxed problem, we can employ the Lagrange multiplier to offer an alternative representation:
{\scriptsize
\begin{equation}
    \underset{\Delta\mathbf{W}}{\mathrm{minimize}} \left( c \cdot \sum_{\mathbf{x}\in D} p(\mathbf{x}, \{f, \mathbf{W}+\Delta\mathbf{W}\}) + (\mathcal{L}(\Delta\mathbf{W}) - th_g)\right)\label{eq:final}
\end{equation}}
given that \( c > 0 \) is a carefully selected constant, the optimal solution is achievable. This objective corresponds to the relaxed problem, in which, for some \( c > 0 \), the optimal solution to the latter matches that of the former.

Consequently, we employ the optimization objective (\ref{eq:final}) as the relaxed alternative to our originally defined objective (\ref{eq:ori_def}). Given that the objective (\ref{eq:final}) is differentiable with respect to $\Delta\mathbf{W}$, we utilize gradient descent as the optimization algorithm to address this issue.

\textbf{Methods to determine constant $c$.}

\begin{figure}[ht]
    \includegraphics[trim=0 140 260 0, clip, width=0.8\linewidth]{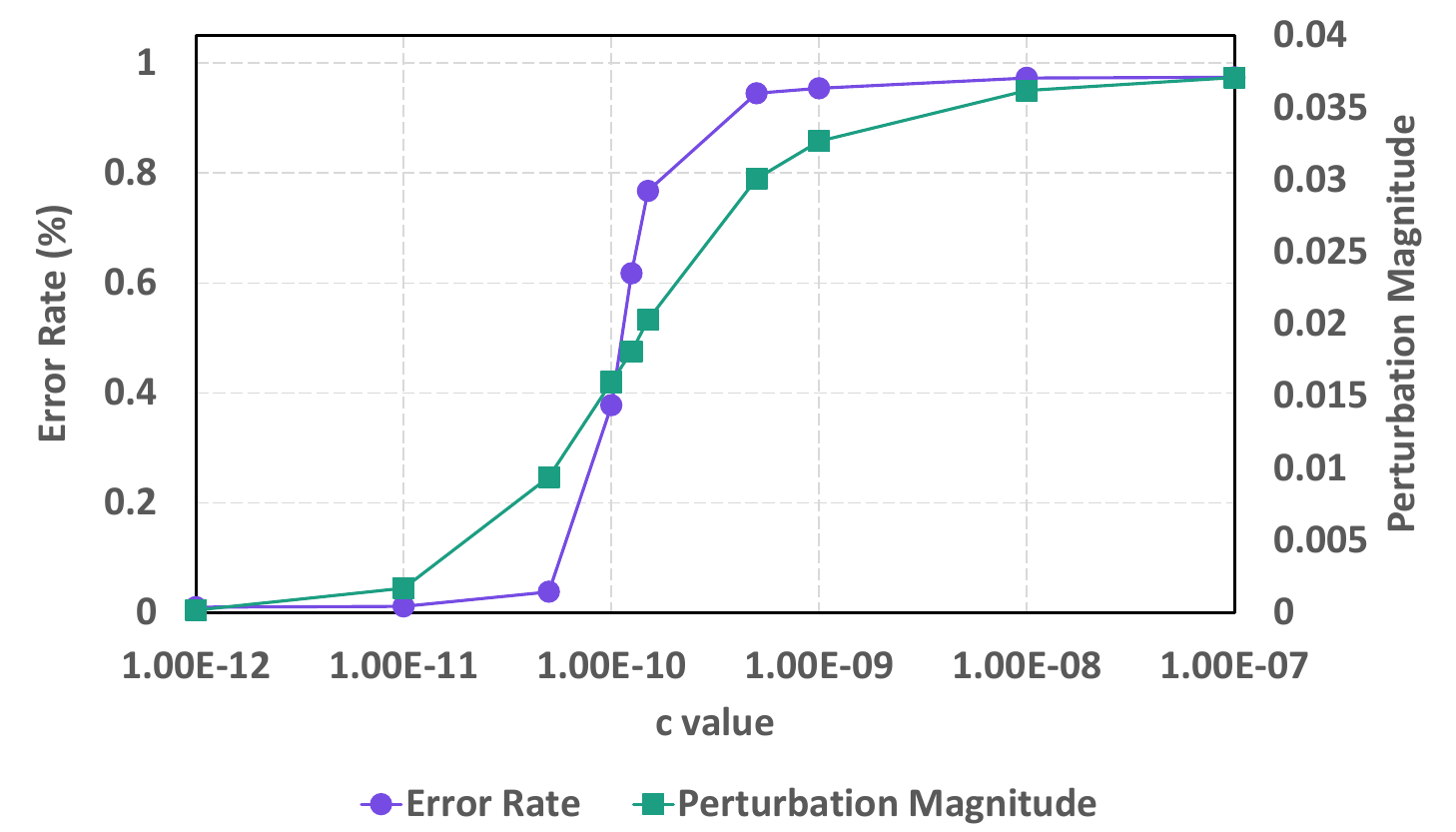}
\caption{
Selection of the constant $c$. We illustrate two distinct relationships: (1) The worst-case error rate (left) as a function of the $c$ value and (2) The perturbation magnitude $\mathcal{L}(\Delta\mathbf{W})$ (right) as a function of the $c$ value. The DNN model under consideration is LeNet applied to the MNIST dataset.
}
\label{fig:CvA}
\end{figure}

In qualitative terms, when examining objective (\ref{eq:final}), larger values of $c$ indicate a heightened emphasis on reducing accuracy and a diminished emphasis on $\mathcal{L}(\Delta\mathbf{W})$. This typically leads to a decreased final accuracy and an increased value of $\mathcal{L}(\Delta\mathbf{W})$. This observation is substantiated by the empirical results illustrated in Fig.~\ref{fig:CvA}. Here, we depict how the worst-case error rate and $\mathcal{L}(\Delta\mathbf{W})$ fluctuate based on the chosen $c$, utilizing LeNet for MNIST as an example. 

Given that empirical outcomes indicate $\mathcal{L}(\Delta\mathbf{W})$ displays monotonic behavior with respect to $c$, our strategy to determine the optimal $c$ value that results in the minimum performance, subject to the weight perturbation constraint $th_g$, is to employ binary search. Our goal is to identify the largest $c$ value that satisfies $\mathcal{L}(\Delta\mathbf{W}) \leq th_g$. The accuracy derived using this particular $c$ value represents the worst-case performance of the DNN model when weight perturbations are confined to $th_g$.

It is worthwhile to note that since the optimization problem outlined in (\ref{eq:final}) is solvable through gradient descent, both the time and memory complexities of our method align closely with those required for training the DNN. This final framework of evaluating the worst-case performance is named Lagrange-based Worst-Case analysis (\LWC).
\begin{table*}[t]
    \centering
    \caption{
    Comparison between the proposed \LWC~method and two baseline techniques: Monte Carlo simulation (MC) and weight PGD attack (PGD), in determining the worst-case accuracy of different DNN models across various datasets, given a weight perturbation bound of $th_g = 0.03$. Also listed is the accuracy of the original model without any perturbation (Ori. Acc). Our method identifies perturbations that result in considerably lower accuracy compared to other techniques. While it takes slightly more time than the weight PGD attack, it is significantly faster than the MC simulation.
    }
    \begin{tabular}{cccrrrrrr}
        \toprule
        \multirow{2}{*}{Dataset} & \multirow{2}{*}{Model}          &\multirow{2}{*}{Ori. Acc.} & \multicolumn{3}{c}{Worst-case Accuracy (\%)} & \multicolumn{3}{c}{Running Time (Minutes)}\\ 
                        &     &    & MC & PGD & \LWC & MC & PGD & \LWC\\
        \midrule
        MNIST           & LeNet     & 99.12 & 97.34  & 13.44 & \textbf{7.35} & 900    & 3.3  & 5.5\\
        CIFAR-10        & ConvNet   & 85.31 & 60.12  & 10.00 & \textbf{4.27} & 2700   & 4.2  & 6.0\\
        CIFAR-10        & ResNet-18 & 95.14 & 88.77  & 10.00 & \textbf{0.00} & 5400   & 13.3 & 20.0\\
        Tiny ImageNet         & ResNet-18 & 65.23 & 25.33  & 0.50 & \textbf{0.00} & 14400  & 40.0 & 60.0\\
        ImageNet        & ResNet-18 & 69.75 & 43.98  & 0.10 & \textbf{0.00} & 231000 & 1980 & 2880\\
        ImageNet        & VGG-16    & 71.59 & 66.43  & 0.10 & \textbf{0.06} & 313800 & 2530 & 3820\\
        \bottomrule
    \end{tabular}
    \label{tab:GPO_res}
\end{table*}

\begin{table}[ht]
    \centering
    \caption{
    Hyper-parameter configurations for executing the proposed \LWC~method across various models and datasets are detailed. This includes the $c$ value determined from (\ref{eq:final}) via binary search, the employed learning rate (lr), and the iteration count for gradient descent.
    }
    \begin{tabular}{cccccc}
        \toprule
        Dataset         & Model     & c    & lr & \# of runs\\
        \midrule
        MNIST           & LeNet     & 1E-3  & 1E-5 & 500\\
        CIFAR-10        & ConvNet   & 1E-5  & 1E-5 & 100\\
        CIFAR-10        & ResNet-18 & 1E-9  & 1E-4 & 20 \\
        Tiny ImgNet     & ResNet-18 & 1E-10 & 1E-4 & 20 \\
        ImageNet        & ResNet-18 & 1E-3  & 1E-3 & 10 \\
        ImageNet        & VGG-16    & 1E-3  & 1E-3 & 10 \\
        \bottomrule
    \end{tabular}
    \label{tab:GPO_setup}
\end{table}

\subsection{Experimental Evaluation}\label{sect:exp}

In this section, we evaluate the efficacy of our proposed \LWC~approach in determining the worst-case performance of various DNN models through experiments. We utilized six distinct DNN models across four datasets: LeNet~\cite{lecun1998gradient} model targeting the MNIST~\cite{deng2012mnist} dataset, ConvNet~\cite{peng2019dnn+} model targeting the CIFAR-10~\cite{krizhevsky2009learning} dataset, ResNet-18~\cite{he2016deep} model targeting the CIFAR-10 dataset, ResNet-18 model targeting the Tiny ImageNet~\cite{le2015tiny} dataset, ResNet-18 model targeting the ImageNet~\cite{deng2009imagenet} dataset, and VGG-16~\cite{simonyan2014very} model targeting the ImageNet dataset.
The inference process of LeNet and ConvNet models use a quantization precision of 4 bits, for both weights and layer inputs. On the other hand, the ResNet-18 and VGG-16 models are quantized to 8 bits. For the purposes of this study, we adopt $K=2$ in line with~\cite{jiang2020device, yan2022swim}. Following the established procedures detailed in Section~\ref{sec:2.3}, for each weight, the discrepancy between the actual device value and the anticipated value is progressively programmed until it falls below 0.1. Specifically, this means $th = 0.06$ as per~\cite{yan2022swim}, resulting in a $th_g$ value of 0.03 (unless stated otherwise). These figures align with those presented in~\cite{shim2020two}, corroborating the authenticity of our modeling and the parameters employed.

Given the absence of prior research for identifying a DNN's worst-case performance under device variations, besides the basic MC simulations, we also utilize a modified version of the weight PGD attack method~\cite{tsai2021formalizing}. This modified method aims to discern the minimal weight perturbation resulting in a successful attack, serving as an auxiliary reference point. All experiments were executed on Titan-XP GPUs using the PyTorch framework. For the MC simulation baseline, we employed 100,000 runs and leveraged the Adam optimizer for gradient descent. The comprehensive setup for our proposed \LWC~method is detailed in Table~\ref{tab:GPO_setup}.

\subsubsection{Worst-case DNN Accuracy Obtained by Different Methods}

Table~\ref{tab:GPO_res} reveals that our proposed \LWC~framework is superior in identifying the worst-case performance when compared to the weight PGD attack and MC simulations. For DNN models like LeNet and ConvNet, our method uncovers weight perturbations that plunge accuracy to below 10\%. For ResNet-18 and VGG-16, the accuracy is almost 0\%. In contrast, the weight PGD attack can only locate perturbations resulting in accuracy close to what one would expect by random guessing (\emph{i.e.}, 1/N for classifying images to N different classes, translating to 10\% in the MNIST and CIFAR-10 dataset, 0.5\% in the Tiny ImageNet dataset, and 0.1\% in the ImageNet dataset). Methods using MC simulations are the least effective, as even after 100,000 runs, they do not approach the accuracy drops achieved by the other techniques. Given the expansive exploration space created by numerous weights, this outcome is unsurprising.

In terms of computation time, our framework requires slightly more time than the weight PGD attack, primarily due to the gradient descent's convergence time. Nevertheless, both methods considerably outpace MC simulations in terms of execution time.

These findings indicate a significant vulnerability of DNNs to device variations, even when using the write-verify method and maintaining a maximum weight perturbation of only $0.03$. Considering that even extensive MC simulations can not indentify the genuine worst-case accuracy, safety-critical applications may necessitate individual inspections of each programmed CiM accelerator to ensure its performance. Relying on random sampling for quality assurance may be too risky.

Furthermore, by comparing our framework's results for ConvNet and ResNet-18 models on the CIFAR-10 dataset (and also VGG-16 and ResNet-18 models on the ImageNet dataset), it becomes evident that deeper networks are more vulnerable to weight perturbations. This trend aligns with the expectation that greater perturbations can accumulate during forward propagation.

Lastly, the experiments underscore that quantization of both weights and activations is not a reliable strategy for bolstering worst-case DNN performance, considering all the models in this study were quantized as specified in the experimental setup.

\subsubsection{Analysis of Classification Results}

\ 

\begin{figure}[ht]
    \centering
    \includegraphics[trim=0 0 0 0, clip, width=0.8\linewidth]{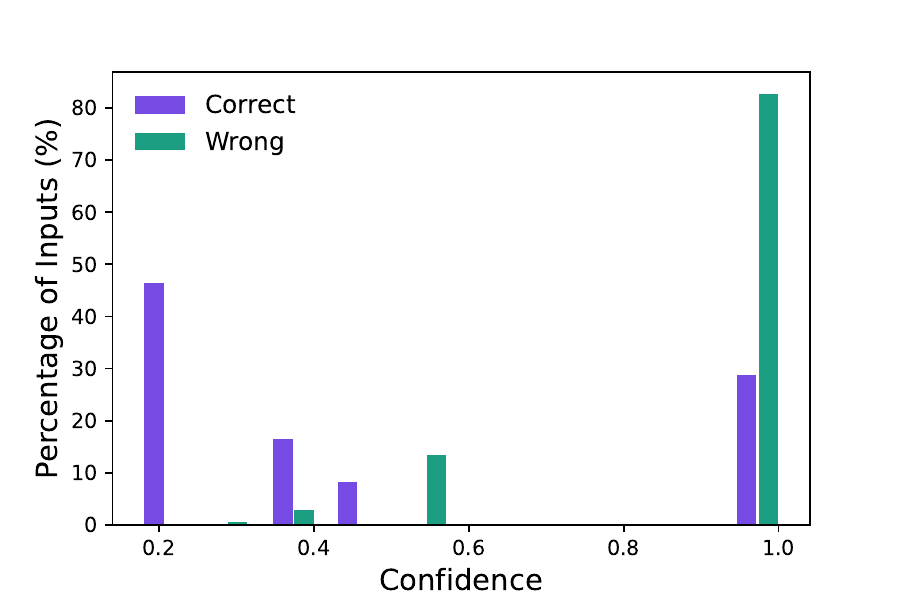}
    \caption{
    Distribution of classification confidence for the worst-case LeNet on MNIST. Surprisingly, the model exhibits greater confidence in its incorrect classifications compared to the correct ones.
    }
    \label{fig:conf_dist}
\end{figure}

Upon closely analyzing the classification outcomes of the worst-case LeNet model identified by our framework for classifying the MNIST dataset, some intriguing observations emerge. A key aspect we scrutinized was classification confidence, whose distribution is illustrated in Fig.~\ref{fig:conf_dist}. Typically, a DNN's classification confidence for one certain input image is gauged using the Softmax function applied to its output vector, with the element showcasing the highest score after the Softmax process deemed as the classification result and this score appointed to be the classification confidence. Surprisingly, the figure indicates that the worst-case LeNet confidently misclassifies all inputs, boasting a 0.90 average confidence score for incorrectly classified inputs. Conversely, for correctly classified inputs, its confidence is notably lower, averaging merely 0.47. This starkly contrasts with the original, unperturbed LeNet, which consistently displays confidence levels nearing 1~\cite{lecun1998gradient}.

Further examination reveals intriguing patterns in classification distribution across different classes, detailed in Table~\ref{tab:class}. The table underscores that a significant chunk of errors stems from images being erroneously categorized into the same class (class 1). Concurrently, a vast majority of images inherently belonging to this class are misclassified into other categories (specifically class 2 and class 3).

We anticipate that these insights will serve as valuable reference points, potentially guiding the creation of innovative algorithms designed to bolster the worst-case performance of DNNs in subsequent studies.

\begin{table}[t]
    \centering
    \caption{
    Normalized classification outcomes for the worst-case LeNet on MNIST. The value at row $i$ and column $j$ represents the proportion of instances with true class $i$ that are incorrectly classified as class $j$, normalized by the total instances of class $i$. Notably, the majority of inputs are incorrectly identified as class 1.
    }
    \begin{tabular}{cc|cccccccccc}
        \toprule
        & &\multicolumn{10}{c}{Model Outputs}\\
        & & 0 & 1 & 2 & 3 & 4 & 5 & 6 & 7 & 8 & 9 \\ 
        \midrule
        \multirow{10}{*}{\rotatebox[origin=c]{90}{Ground Truth Labels}}
        &0& 0.0 & \red{1.0} & 0.0 & 0.0 & 0.0 & 0.0 & 0.0 & 0.0 & 0.0 & 0.0 \\ 
        &1& 0.0 & 0.4 & \red{0.5} & \red{0.1} & 0.0 & 0.0 & 0.0 & 0.0 & 0.0 & 0.0 \\ 
        &2& 0.0 & \red{1.0} & 0.0 & 0.0 & 0.0 & 0.0 & 0.0 & 0.0 & 0.0 & 0.0 \\ 
        &3& 0.0 & \red{1.0} & 0.0 & 0.0 & 0.0 & 0.0 & 0.0 & 0.0 & 0.0 & 0.0 \\ 
        &4& 0.0 & \red{0.9} & 0.0 & 0.0 & 0.0 & 0.0 & 0.0 & \red{0.1} & 0.0 & 0.0 \\ 
        &5& 0.0 & \red{0.9} & 0.0 & \red{0.1} & 0.0 & 0.0 & 0.0 & 0.0 & 0.0 & 0.0 \\ 
        &6& 0.0 & \red{1.0} & 0.0 & 0.0 & 0.0 & 0.0 & 0.0 & 0.0 & 0.0 & 0.0 \\ 
        &7& 0.0 & \red{0.6} & 0.0 & \red{0.4} & 0.0 & 0.0 & 0.0 & 0.0 & 0.0 & 0.0 \\ 
        &8& 0.0 & \red{0.8} & 0.0 & \red{0.2} & 0.0 & 0.0 & 0.0 & 0.0 & 0.0 & 0.0 \\ 
        &9& 0.0 & \red{0.9} & 0.0 & \red{0.1} & 0.0 & 0.0 & 0.0 & 0.0 & 0.0 & 0.0 \\ 
        \bottomrule
    \end{tabular}
    \label{tab:class}
\end{table}

\subsubsection{Distribution of Worst-Case Weight Perturbation}

\ 

\begin{figure}[ht]
    \centering
    \includegraphics[trim=0 0 0 0, clip, width=0.8\linewidth]{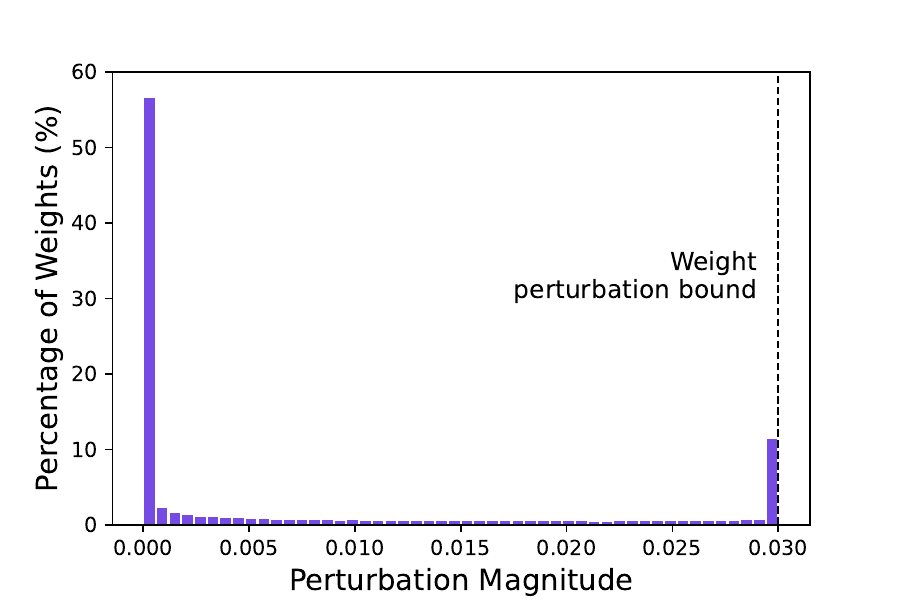}
    \caption{
    Distribution of perturbation magnitude across weights in the worst-case LeNet for MNIST. A significant number of weights remain unperturbed, while many others experience perturbations up to the limit of $th_g$.
    }
    \label{fig:mag_dist}
\end{figure}

In this section, we present the distribution of perturbations among the weights. Here we demonstrate an example using the worst-case LeNet model for the MNIST dataset. As illustrated in Fig.~\ref{fig:mag_dist}, a majority (65\%) of the weights are either left unperturbed or perturbed to their maximum possible extent, that is, $th_g=0.03$. 

\begin{figure}[ht]
    \centering
    \includegraphics[trim=0 110 300 0, clip, width=0.7\linewidth]{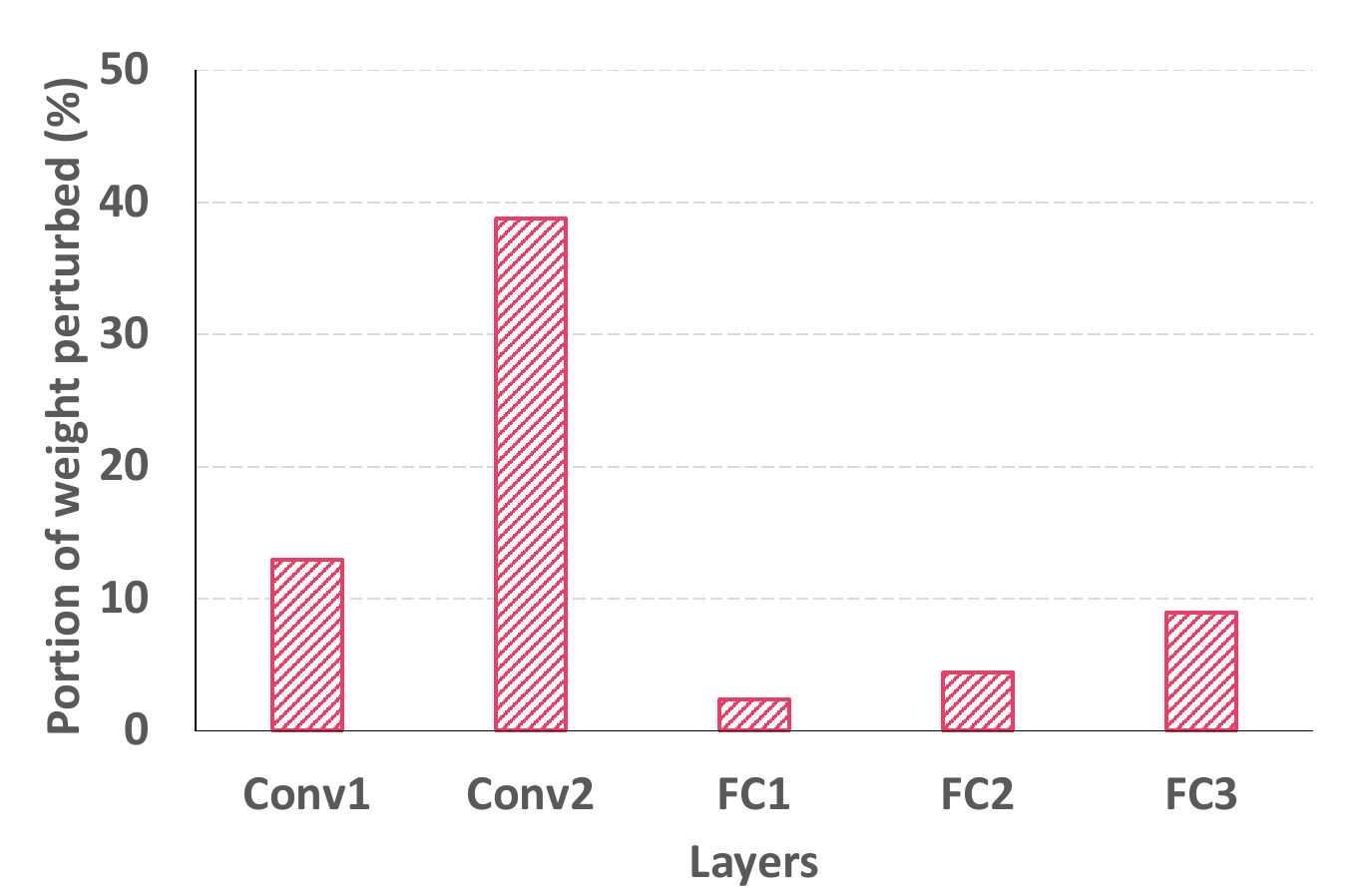}
    \caption{
    Distribution of perturbed weights by layer in the worst-case LeNet for MNIST. Convolutional layers and the final FC layer exhibit a higher percentage of perturbed weights.
    }
    \label{fig:layer}
\end{figure}

Further, in Fig.~\ref{fig:layer}, we detail the count of perturbed weights across different layers. A notable observation is that weights in the convolutional layers, as well as the final fully connected (FC) layer, are more frequently perturbed. This likely stems from their significant influence on the overall accuracy of the DNN.

\section{Enhancing Worst-Case Performance of CiM DNN Accelerators}\label{sect:protect}
Several strategies have been proposed in existing literature aiming to enhance the average performance of DNNs under the influence of device variations. In this section, we aim to adapt these strategies to optimize the worst-case performance of DNNs and evaluate their efficacy. We will explore two main categories of methods: (1) Restricting device variations and (2) Training DNNs to be more resilient to device variations.
From our discussions in Section~\ref{sect:related}, write-verify emerges as a popular strategy in the first category. For the second approach, variation-aware training is commonly adopted. 
\todo{
Before assessing the effectiveness of these methods, we introduce a novel evaluation technique designed to efficiently estimate the worst-case performance of DNN models under the influence of device variations. Utilizing this rapid evaluation method, we then appraise the efficacy of the existing approaches and demonstrate that they are either ineffective or overly expensive. As a result, they are unsuitable for enhancing worst-case performance. Based on these observations, we further propose incorporating a novel training method, TRICE~\cite{yan2023improving}, in conjunction with the adversarial training approach~\cite{tsai2021formalizing}, to improve the worst-case performance. The combined worst-case-aware training method is named A-TRICE.
}

Unless otherwise indicated, all accuracy metrics presented in this section are derived from training a single DNN architecture with consistent specifications but using \textbf{three} distinct initializations. The accuracy (and error rate) figures are presented as percentages in the format [average $\pm$ standard deviation] for these three iterations. \todo{
Note that although we use \FLWC~to find the optimal hyper-parameters for each protection method, the final worst-case performance results are generated by the more precise \LWC~method.
}

\subsection{Fast Evaluation Methods}
\todo{
While we can determine the worst-case performance of DNN models under the influence of device variations by using \LWC, it still requires multiple gradient descent trials to identify the Lagrange multiplier constant \( c \). This introduces a considerable time overhead. Therefore, \LWC~is not cost-effective for evaluating the effectiveness of different protection methods, especially since this evaluation necessitates extensive hyper-parameter tuning. Each hyper-parameter selection must be assessed independently, so a swifter and more adaptable approach is preferred.
To address this challenge, we introduce a relaxed version of \LWC~to estimate the worst-case performance. Instead of employing the solution from Eq.~\ref{eq:final} for the constrained optimization problem in Eq.~\ref{eq:rel_problem}, we use a different approach. While \LWC~solves the optimization problem in Eq.~\ref{eq:rel_problem} through the Lagrange multiplier, our proposed rapid evaluation technique employs the fast gradient sign method (FGSM) for this constrained optimization.
}

\todo{
Specifically, to minimize Eq.~\ref{eq:rel_problem} under the constraint \( \mathcal{L}(\Delta\mathbf{W}) \leq th_g \), we can start with \( \Delta\mathbf{W} \) set to zero and continuously increase (or decrease) \( \Delta\mathbf{W} \) by a fixed step size $\eta$ until \( \mathcal{L}(\Delta\mathbf{W}) \) reaches \( th_g \). Because $\Delta\mathbf{W}$ is relatively small, the convergence is typically reached in $th_g/\eta$ iterations. By controlling $\eta$, a trade-off between execution time and precision can be achieved. The worst-case accuracy of the DNN model is then defined by the accuracy of the model with weights equal to \( \mathbf{W} + \Delta\mathbf{W} \). We term this method \FLWC. Further details of this approach can be found in Algorithm~\ref{alg:flwc}.
}

\begin{algorithm}[ht]
\caption{F-LWC~($f$, $W$, $D$, $Ep$, $p$, $th_g$, $\eta$)}
\begin{algorithmic}[1]\label{alg:flwc}
\STATE // INPUT: A DNN architecture $f$, DNN weight $\mathbf{W}$, training dataset $\mathbf{D}$, loss function $l$ which is Eq.~\ref{eq:rel_ch}, perturbation distance $th_g$, step size $\eta$;
\STATE // Output: Weight perturbation $\Delta\mathbf{W}$;
\STATE Initialize weight perturbation $\Delta\mathbf{W}$ to be a vector of all zero and of the same size as $\mathbf{W}$;
\STATE $Ep = th_g / \eta$
\FOR{($i=0$; $i < Ep$; $i++$)}
    \FOR{mini-batches $\mathbf{B}$ in $\mathbf{D}$}
        \STATE Divide $\mathbf{B}$ into input $\mathbf{I}$ and label $\mathbf{L}$;
        \STATE $\mathbf{O} = f(\mathbf{W} + \Delta\mathbf{W}, \mathbf{I})$;
        \STATE $loss_p = p(\mathbf{O}, \mathbf{L})$;
        \STATE $\Delta\mathbf{W} = \Delta\mathbf{W} - \eta \times \text{sign}(\frac{\partial loss_p}{\partial \mathbf{W}})$
        
    \ENDFOR
\ENDFOR
\STATE Return $\Delta\mathbf{W}$
\end{algorithmic}
\end{algorithm}

\begin{table}[t]
    \centering
    \caption{
    \todo{
    Comparison between the proposed fast evaluation method and the original worst-case evaluation method in determining the worst-case accuracy of different DNN models across various datasets, given a weight perturbation bound of $th_g = 0.03$. Also listed is the accuracy of the original model without any perturbation (Ori. Acc).
    }
    }
    \begin{tabular}{cccrrrr}
        \toprule
        \multirow{2}{*}{Dataset} & \multirow{2}{*}{Model}          &Ori. & \multicolumn{2}{c}{WC Acc. (\%)} & \multicolumn{2}{c}{Time (Min)}\\ 
                        &           & Acc   & \LWC & \FLWC  & \LWC & \FLWC\\
        \midrule
        MNIST           & LeNet     & 99.12 & \textbf{7.35} & 11.31 & 5.5  &  2.1\\
        CIFAR-10        & ConvNet   & 85.31 & 4.27 &  \textbf{1.51} & 6.0  &  3.0\\
        CIFAR-10        & ResNet-18 & 95.14 & \textbf{0.00} &  2.03 & 20.0 &  9.5\\
        Tiny ImageNet   & ResNet-18 & 65.23 & \textbf{0.00} &  0.00 & 60.0 & 28.7\\
        \bottomrule
    \end{tabular}
    \label{tab:FE_res}
\end{table}

\todo{
We evaluate the efficacy of \FLWC~by contrasting it with \LWC. Given that the weight perturbation distance adheres strictly to the constraint \( th_g \), a lower worst-case accuracy implies a more accurate evaluation method. As illustrated in Table~\ref{tab:FE_res}, both \FLWC~and~\LWC are used to assess the worst-case performance of four distinct DNN models. From an efficiency standpoint, \FLWC~boasts a 2x speedup relative to \LWC. Regarding accuracy, \FLWC~identifies a worst-case scenario closely aligning with that identified by \LWC~in three models: the LeNet model targeting the MNIST dataset, the ResNet-18 model targeting the CIFAR-10 dataset, and the ResNet-18 model targeting the Tiny ImageNet dataset. Intriguingly, for the ConvNet model for CIFAR-10, \FLWC~determines an even lower accuracy than \LWC. This discrepancy is rational. \LWC~demands considerable hyper-parameter tuning, and the values reported in our prior work, which initially introduced \LWC~\cite{yan2022computing}, may not represent the optimal outcome. However, we maintain this value for consistency with the reported results. Conversely, \FLWC~demands minimal hyper-parameter tuning, making it simpler to pinpoint its optimal outcome.
}

\begin{figure}[ht]
    \centering
    \includegraphics[trim=0 0 0 0, clip, width=0.8\linewidth]{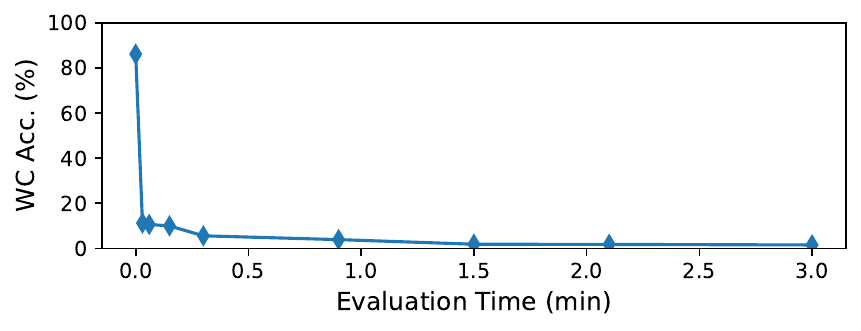}
    \caption{
    The time/precision trade-off for \FLWC. Experiment done on ConvNet for CIFAR-10 with $th_g = 0.03$.
    }
    \label{fig:tradeoff}
\end{figure}

\todo{
Further illustrating \FLWC's adaptability, we augment its step size. A larger step size equates to reduced runtime but less accurate estimation, thus presenting a trade-off between these aspects. In Fig.~\ref{fig:tradeoff}, we highlight that, even within a limited runtime, \FLWC~still provides a reasonably accurate estimation.
}

\subsection{Conventional Methods for Improving DNN Robustness}~\label{sect:bad}
\todo{
In this section, we demonstrate that all of the existing methods used to improve the nvCiM DNN accelerator robustness are not effective in improving the worst-case performance of DNN models. We first discuss the existing conventional methods and then demonstrate their effectiveness.
}

\subsubsection{Stronger Write-Verify}\label{sect:wv}

As indicated in Table~\ref{tab:GPO_res}, merely relying on the conventional write-verify procedure to restrict the maximum weight perturbation to 0.03 (as specified by $th_g = 0.03$ in (\ref{eq:noise})) does not substantially boost the worst-case performance of DNN models. By designating a smaller $th_g$ for the write-verify process, the writing duration may increase. However, this could potentially enhance the DNN's worst-case performance. More analysis are shown in Fig.~\ref{fig:all_ORI}.

\subsubsection{Variation-Aware Training}\label{sect:vt}
We have discussed the idea and details of the noise-injection training-based variation-aware training method, where Gaussian noises are injected in the forward and backward process of DNN training to improve the DNN robustness.

\subsubsection{Adversarial Training}\label{sect:tm}

\begin{algorithm}[t]
\caption{Adversarial Training~($f$,  $D$, $\mathbf{V}$, $Ep$, $l$, $\eta$)}
\begin{algorithmic}[1]\label{alg:adv}
\STATE // INPUT: A DNN architecture $f$, training dataset $\mathbf{D}$, validation dataset $\mathbf{V}$, the total number of training epochs $Ep$, loss function $l$, learning rate $\eta$;

\STATE Initialize weight $\mathbf{W}$ for $f$;
\STATE Initialize $acc_F = 0$, $\mathbf{W}_F = \mathbf{W}$;

\FOR{($i=0$; $i < Ep$; $i++$)}

    \FOR{mini-batches $\mathbf{B}$ in $\mathbf{D}$}
        \STATE Divide $\mathbf{B}$ into input $\mathbf{I}$ and label $\mathbf{L}$;
        \STATE Find weight perturbations $\mathbf{N}$ that lead to worst-case accuracy using the framework discussed in Section~\ref{sect:method};
        \STATE $\mathbf{O} = f(\mathbf{W} + \mathbf{N}, \mathbf{I})$;
        \STATE $loss = l(\mathbf{O}, \mathbf{L})$;
        \STATE $\mathbf{G} = \frac{\partial loss}{\partial \mathbf{W}}$;
        \STATE $\mathbf{W} = \mathbf{W} - \eta \times \mathbf{G}$
        
    \ENDFOR
    \STATE Evaluate $\mathbf{W}$ on $\mathbf{V}$ and get $acc$;
    \IF{$acc > acc_F$}
        \STATE $acc_F = acc$;
        \STATE $\mathbf{W_F} = \mathbf{W}$
    \ENDIF
\ENDFOR
\STATE Return $\mathbf{W}_F$
\end{algorithmic}
\end{algorithm}

\begin{table}[t]
    \centering
    \caption{
    Table displaying the worst-case accuracy (in \%) of diverse DNN models under regular training (Regular), variation-aware training (VA), and adversarial training (ADV). The write-verify process is conducted with a weight perturbation boundary set at $th_g = 0.03$. While adversarial training shows efficacy with the LeNet model for MNIST, neither method appears effective for other, more intricate models.
    }
    \begin{tabular}{ccrrr}
        \toprule
        Dataset         & Model     & Regular    & VA & ADV \\
        \midrule
        MNIST           & LeNet     & 7.35$\pm$03.70 & 81.58$\pm$00.80 & 98.26$\pm$01.05\\
        CIFAR10         & ConvNet   & 4.27$\pm$00.33 &  3.71$\pm$ 3.76 & 67.09$\pm$03.85\\
        CIFAR10         & ResNet18  & 0.00$\pm$00.00 &  2.84$\pm$ 1.20 & 34.84$\pm$13.20\\
        Tiny IN         & ResNet18  & 0.00$\pm$00.00 & 3.57$\pm$03.48 & 7.41$\pm$08.10\\
        \bottomrule
    \end{tabular}
    \label{tab:protect}
\end{table}

Here we introduce a modified version of the adversarial training method~\cite{tsai2021formalizing}. Adversarial training is traditionally employed to counteract adversarial input, as summarized in Alg.~\ref{alg:adv}. Drawing parallels to the way adversarial training addresses input perturbations, we inject worst-case perturbations into the weights of a DNN during the training process, aiming to enhance its performance amidst device variations. Specifically, during each training iteration, we first employ \LWC~to identify the perturbations on the current weights that result in the model's worst-case accuracy. Subsequently, we integrate these perturbations into the weights and gather the gradient $G$.

Table~\ref{tab:protect} demonstrates that both variation-aware training and adversarial training generally yield improvements over regular training. Adversarial training exhibits a slightly superior performance in comparison to variation-aware training. Nevertheless, when benchmarked against the optimal accuracy achievable by these networks in the absence of device variations (refer to the third column in Table~\ref{tab:GPO_res}), there remains a notable decline in worst-case accuracy. A notable exception is observed in the LeNet model for MNIST. In this model, adversarial training almost entirely compensates for the accuracy reduction even under worst-case scenarios, a result attributable to the model's simplicity. Furthermore, as the complexity of the network increases, the enhancements offered by both training methods appear to wane, as evidenced by the mere $7.41\%$ improvement for ResNet-18 on the Tiny ImageNet dataset.

\begin{figure*}
\subfigure[LeNet for MNIST]{
    \includegraphics[trim=10 0 30 30, clip, width=0.32\linewidth]{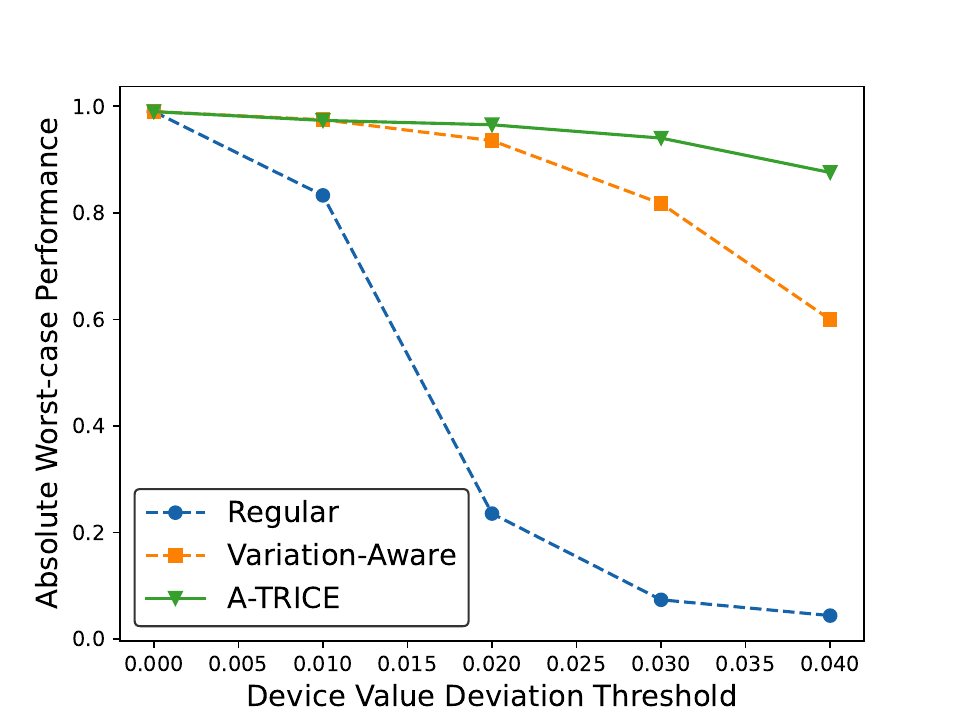}
    \label{fig:diff_dist_lenet}}
\subfigure[ConvNet for CIFAR-10]{
    \includegraphics[trim=8 0 32 30, clip, width=0.32\linewidth]{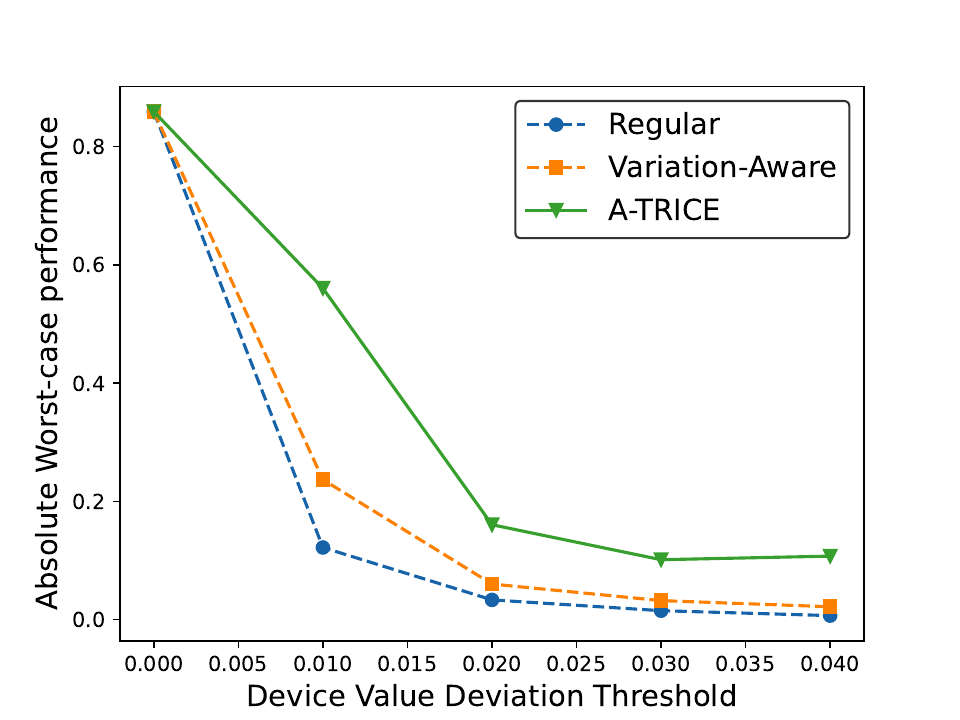}    
    \label{fig:diff_dist_cifar}}
\subfigure[ResNet18 for CIFAR-10]{
    \includegraphics[trim=10 0 30 30, clip, width=0.32\linewidth]{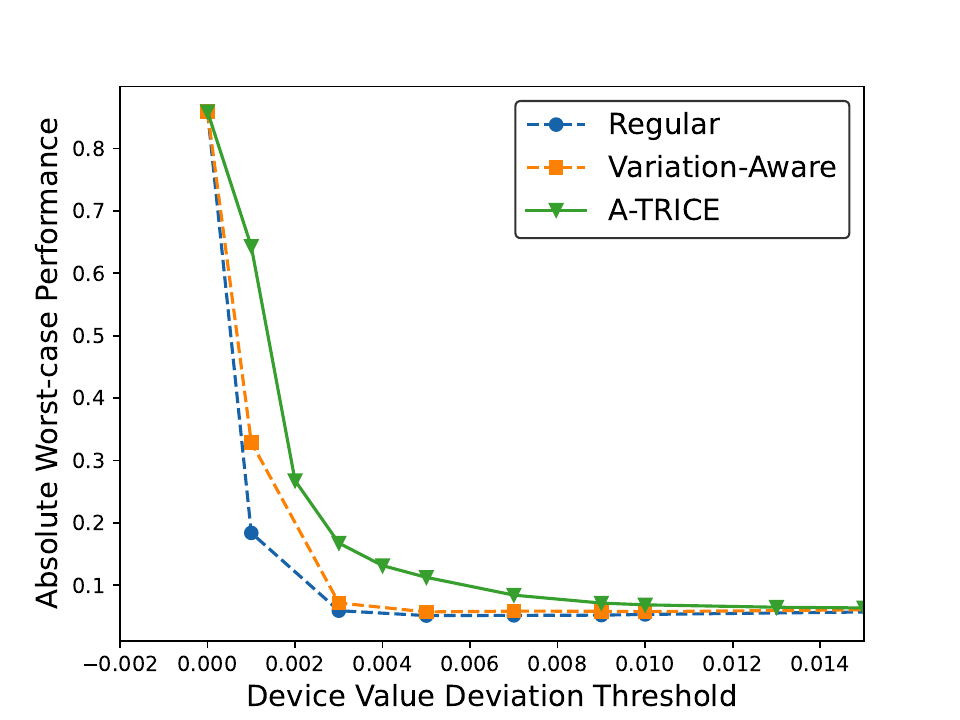}
    \label{fig:diff_dist_res18}}
\caption{
\todo{
Examining the effectiveness of stronger write-verify combined with the proposed A-TRICE method (green triangles). The baseline methods are regular training (blue dashed line) and variation-aware training (orange dashed line). Each figure depicts the correlation between the weight perturbation limit during write-verify, denoted as $th_g$ (X-axis), and the worst-case accuracy of the DNN (Y-axis) for various models: (a) LeNet for MNIST, (b) ConvNet for CIFAR-10, and (c) ResNet-18 for CIFAR-10. For every data point, five experiments with identical settings but varied random initializations are performed.
}
}
\label{fig:all_ORI}
\end{figure*}

\todo{
The blue dashed curve in Fig.~\ref{fig:all_ORI} shows the effectiveness of stronger write-verify. Stronger write-verify can be effective when the write-verify is done with very strong effort (can provide a very tight bound in device value). Specifically, stronger write-verify can preserve high enough accuracy for the LeNet model when $th_g = 0.01$, which would require significant programming time. Stronger write-verify is not effective for larger models like ConvNet and ResNet-18 even with this threshold value. Note that the maximum $th_g$ shown in the figure is $0.04$, which is greater than previous experiments. This can be viewed as a weaker write-verify scheme.
}

\subsection{A-TRICE Algorithm}

\todo{
As demonstrated in Sect.~\ref{sect:bad}, conventional methods are not effective in improving the worst-case performance of DNN models under the influence of device variations. Here we further propose a novel training method named \underline{A}dversarial \underline{T}raining with \underline{RI}ght-\underline{C}ensored Gaussian Nois\underline{E} (A-TRICE).}

\todo{
\subsubsection{A-TRICE Method}
To improve the worst-case performance, we resort to our recent work TRICE~\cite{yan2023improving} which can effectively improve the quantile performance of DNN models. The key to this method is to inject a novel type of noise named Right-Censored Gaussian (RCG) noise instead of the widely used Gaussian noise in the training process. In an RCG distribution, all values follow the Gaussian distribution except that those greater than a certain threshold are set (censored) to be the threshold value. Formally, we have:
\begin{equation}
    \begin{aligned}
        \text{\emph{RC-Gaussian}}(th, \sigma_t) &= \begin{cases}
                th \times \sigma_t,   &\emph{if} \  g \geq th \times \sigma_t \\
                g,   &\emph{else}
            \end{cases}& \\
        g &\sim \mathcal{N}(0, \sigma_t) 
    \end{aligned}\label{eq:cr-gaussian}
\end{equation}
where $th$ is the use-defined cutoff value and $\sigma_t$ is the standard deviation of the corresponding Gaussian distribution.
Although TRICE was originally proposed to improve the percentile performance of DNN models, it is also effective when the percentile is small, mimicking a worst-case scenario.
}

\todo{
We incorporate a novel efficient adversarial training method into the TRICE framework so that the resultant A-TRICE framework can efficiently improve the worst-case performance of DNN models.
A-TRICE aggregates the adversarial training and noise injection training in one training pass and balances the contribution of these two methods using a coefficient when summing the loss function of the two methods. Specifically, in each iteration of training, A-TRICE first uses the batched input data to approximate a model that would result in the worst-case accuracy using \FLWC. It then performs an adversarial training pass using the worst-case model and collects a resultant $loss_p$. After that, A-TRICE samples a noise instance from RCG distribution, injects this noise into the training process, and collects a resultant $loss_n$. The two resultant losses are then combined using a weighted sum and A-TRICE then use this sum to update the weights through gradient descent. The detailed algorithm is shown in Alg.~\ref{alg:atrice}.
}

\begin{algorithm}[ht]
\caption{A-TRICE~($f$,  $D$, $\mathbf{V}$, $\mathcal{N}$, $Ep$, $l$, $p$, $s$, $\alpha$, $\eta$)}
\begin{algorithmic}[1]\label{alg:atrice}
\STATE // INPUT: A DNN architecture $f$, training dataset $\mathbf{D}$, validation dataset $\mathbf{V}$, noise distribution $\mathcal{N}$, the total number of training epochs $Ep$, loss function $l$, perturbation function $p$, which is Eq.~\ref{eq:rel_ch}, step size $s$ for perturbation, learning rate $\eta$;

\STATE Initialize weight $\mathbf{W}$ for $f$;
\STATE Initialize $acc_F = 0$, $\mathbf{W}_F = \mathbf{W}$;

\FOR{($i=0$; $i < Ep$; $i++$)}

    \FOR{mini-batches $\mathbf{B}$ in $\mathbf{D}$}
        \STATE Divide $\mathbf{B}$ into input $\mathbf{I}$ and label $\mathbf{L}$;
        \STATE // Adversarial Training;
        \STATE $\mathbf{O_c} = f(\mathbf{W}, \mathbf{I})$;
        \STATE $cost_p = p(\mathbf{O_c}, \mathbf{L})$
        \STATE $\mathbf{O_p} = f\left(\mathbf{W} - s \times \text{sign}\left( \frac{\partial cost_p}{\partial \mathbf{W}}\right), \mathbf{I}\right)$;
        \STATE $loss_p = l(\mathbf{O_p}, \mathbf{L})$;
        \STATE // Noise-injection training;
        \STATE Sample $\Delta \mathbf{W}$ from $\mathcal{N}$;
        \STATE $\mathbf{O_n} = f(\mathbf{W} + \Delta \mathbf{W}, \mathbf{I})$;
        \STATE $loss_n = l(\mathbf{O_n}, \mathbf{L})$;
        \STATE $loss = (1-\alpha)\times loss_n + \alpha \times loss_q$;
        \STATE $\mathbf{W} = \mathbf{W} - \eta \times \frac{\partial loss}{\partial \mathbf{W}}$
        
    \ENDFOR
    \STATE Evaluate $\mathbf{W}$ on $\mathbf{V}$ and get $acc$;
    \IF{$acc > acc_F$}
        \STATE $acc_F = acc$;
        \STATE $\mathbf{W_F} = \mathbf{W}$
    \ENDIF
\ENDFOR
\STATE Return $\mathbf{W}_F$
\end{algorithmic}
\end{algorithm}

\subsubsection{Effectiveness of A-TRICE} 
\todo{
Here we demonstrate the effectiveness of A-TRICE by comparing it with two baseline methods: vanilla DNN training and variation-aware training that injects Gaussian noise into training. We did not compare A-TRICE with adversarial training because adversarial training is a special case of A-TRICE when the trade-off coefficient $\alpha = 1$.
}

\todo{
The comparison result is shown in Fig.~\ref{fig:all_ORI}. Specifically, in Fig.~\ref{fig:diff_dist_lenet}, we compare A-TRICE with baseline methods in improving the worst-case performance of the LeNet model targeting the MNIST dataset. This experiment shows that variation-aware training is already effective in improving the worst-case performance of this model if provided a tight write-verify bound ($th_g \leq 0.02$). A-TRICE further improves the worst-case accuracy by 30\% when $th_g$ is large. Fig.~\ref{fig:diff_dist_cifar} demonstrates the comparison between A-TRICE and baseline methods for model ConvNet targeting the CIFAR-10 dataset. We observe that variation-aware training is not effective in improving the worst-case performance of this model, only improving the accuracy by up to 10\%. A-TRICE further improves the worst-case accuracy by up to 33\%. Finally, we compare A-TRICE with baseline methods in improving the worst-case performance of the ResNet-18 model targeting the CIFAR-10 dataset. The result is shown in Fig.~\ref{fig:diff_dist_res18}. We observe that all methods are not fully effective in improving the worst-case performance when $th_g$ is large, but the variation-aware training method is able to improve the worst-case performance by up to 15\%, compared with the vanilla training method, and A-TRICE further improves the worst-case performance by up to 31\%. 
}

\subsubsection{Ablation study} 
\todo{
To further evaluate the contribution of introducing adversarial training into the TRICE framework, we perform an ablation study that compares A-TRICE with our previously proposed TRICE framework. We compare A-TRICE with TRICE in improving the worst-case performance of the ConvNet model targeting the CIFAR-10 model in Table~\ref{tab:GvPT}. We can observe that TRICE and our proposed A-TRICE are both effective in improving the worst-case performance of the ConvNet model. TRICE improves the worst-case accuracy by up to 22\% compared with the vanilla training method. Although not shown in the table, this also means a 12\% improvement compared with variation-aware training. A-TRICE is even more effective, further improving the worst-case accuracy by up to 21\%.
}

\subsubsection{Discussions}
\todo{
In the previous experiments, we demonstrate that A-TRICE is far more effective than all other alternatives and is thus the most promising candidate for improving the worst-case performance. However, we can also observe that even with A-TRICE, the worst-case performance can still be very low for some models when the write-verify bound is reasonably large (\emph{e.g.}, $th_g = 0.03$). Thus, finding more effective methods to address the worst-case performance issue is still an open problem. We also observe that some DNN architectures are more prone to be affected by device variations. For example, when using the same dataset, ResNets have much lower worst-case performance than plain CNNs. We advocate that designers proceed with caution when using ResNets for safety-critical issues.
}

\begin{table}[ht]
    \centering
    \caption{Ablation Study. Comparing the effectiveness of TRICE and A-TRICE in improving the worst-case performance of ConvNet mode targeting the CIFAR-10 dataset. }
    \begin{tabular}{crrr}
        \toprule
        Dev. var.   &\multicolumn{3}{c}{Training Method}\\
        ($\sigma_d$)& w/o noise & TRICE & A-TRICE \\
        \midrule
        0.00        & 85.93     & 84.52 & 84.17\\
        0.10        & 12.20     & 34.16 & 56.05\\
        0.20        &  3.34     &  8.86 & 16.05\\
        0.30        &  1.51     &  4.30 & 10.13\\
        0.40        &  0.68     &  2.85 & 10.74\\
        \bottomrule
    \end{tabular}
    \label{tab:GvPT}
\end{table}
\section{Conclusions}\label{sect:conclusion}
In this paper, instead of focusing on the average performance of DNNs under device variations in CiM accelerators, we introduce an efficient framework to evaluate their worst-case performance, a crucial metric for safety-critical applications. Through our study, we reveal that even with tightly bounded weight perturbations post write-verify, the accuracy of a well-trained DNN can plummet dramatically. Consequently, caution is advised when implementing CiM accelerators in safety-critical scenarios. For instance, during chip quality control after production, it may be prudent to verify the accuracy of each chip individually instead of relying on random sampling. We further evaluate the current techniques aiming at boosting the average DNN performance in CiM accelerators. \todo{While doing so, we further propose a fast estimation method of the worst-case scenario to evaluate the influence of using different hyper-parameters for certain robustness-improving methods.} Our experiments demonstrate that these methods either impose high costs (as in the case of stronger write-verify) or are ineffective (as seen with training-based methods) when adapted to improve worst-case performance. \todo{Thus, we further introduce a novel training method that combines adversarial training and training with right-censored Gaussian noise. Our proposed method significantly improves the worst-case performance of CiM accelerators.}

\section{Acknowledgement}
This project is partially supported by NSF under grants CCF-1919167, CNS-1822099 and CCF-2028879.

\bibliographystyle{IEEEtran}
\bibliography{M7_References}

\begin{thebibliography}{10}
\providecommand{\url}[1]{#1}
\csname url@samestyle\endcsname
\providecommand{\newblock}{\relax}
\providecommand{\bibinfo}[2]{#2}
\providecommand{\BIBentrySTDinterwordspacing}{\spaceskip=0pt\relax}
\providecommand{\BIBentryALTinterwordstretchfactor}{4}
\providecommand{\BIBentryALTinterwordspacing}{\spaceskip=\fontdimen2\font plus
\BIBentryALTinterwordstretchfactor\fontdimen3\font minus \fontdimen4\font\relax}
\providecommand{\BIBforeignlanguage}[2]{{%
\expandafter\ifx\csname l@#1\endcsname\relax
\typeout{** WARNING: IEEEtran.bst: No hyphenation pattern has been}%
\typeout{** loaded for the language `#1'. Using the pattern for}%
\typeout{** the default language instead.}%
\else
\language=\csname l@#1\endcsname
\fi
#2}}
\providecommand{\BIBdecl}{\relax}
\BIBdecl

\bibitem{yang2020co}
L.~Yang, Z.~Yan, M.~Li, H.~Kwon, L.~Lai, T.~Krishna, V.~Chandra, W.~Jiang, and Y.~Shi, ``Co-exploration of neural architectures and heterogeneous asic accelerator designs targeting multiple tasks,'' in \emph{2020 57th ACM/IEEE Design Automation Conference (DAC)}.\hskip 1em plus 0.5em minus 0.4em\relax IEEE, 2020, pp. 1--6.

\bibitem{sheng2022larger}
Y.~Sheng, J.~Yang, Y.~Wu, K.~Mao, Y.~Shi, J.~Hu, W.~Jiang, and L.~Yang, ``The larger the fairer? small neural networks can achieve fairness for edge devices,'' 2022.

\bibitem{shafiee2016isaac}
A.~Shafiee, A.~Nag, N.~Muralimanohar, R.~Balasubramonian, J.~P. Strachan, M.~Hu, R.~S. Williams, and V.~Srikumar, ``Isaac: A convolutional neural network accelerator with in-situ analog arithmetic in crossbars,'' \emph{ACM SIGARCH Computer Architecture News}, vol.~44, no.~3, pp. 14--26, 2016.

\bibitem{sze2017efficient}
V.~Sze, Y.-H. Chen, T.-J. Yang, and J.~S. Emer, ``Efficient processing of deep neural networks: A tutorial and survey,'' \emph{Proceedings of the IEEE}, vol. 105, no.~12, pp. 2295--2329, 2017.

\bibitem{chen2016eyeriss}
Y.-H. Chen, J.~Emer, and V.~Sze, ``Eyeriss: A spatial architecture for energy-efficient dataflow for convolutional neural networks,'' \emph{ACM SIGARCH computer architecture news}, vol.~44, no.~3, pp. 367--379, 2016.

\bibitem{peng2019dnn+}
X.~Peng, S.~Huang, Y.~Luo, X.~Sun, and S.~Yu, ``Dnn+ neurosim: An end-to-end benchmarking framework for compute-in-memory accelerators with versatile device technologies,'' in \emph{2019 IEEE international electron devices meeting (IEDM)}.\hskip 1em plus 0.5em minus 0.4em\relax IEEE, 2019, pp. 32--5.

\bibitem{yan2020single}
Z.~Yan, Y.~Shi, W.~Liao, M.~Hashimoto, X.~Zhou, and C.~Zhuo, ``When single event upset meets deep neural networks: Observations, explorations, and remedies,'' in \emph{2020 25th Asia and South Pacific Design Automation Conference (ASP-DAC)}.\hskip 1em plus 0.5em minus 0.4em\relax IEEE, 2020, pp. 163--168.

\bibitem{jin2020improving}
S.~Jin, S.~Pei, and Y.~Wang, ``On improving fault tolerance of memristor crossbar based neural network designs by target sparsifying,'' in \emph{2020 Design, Automation \& Test in Europe Conference \& Exhibition (DATE)}.\hskip 1em plus 0.5em minus 0.4em\relax IEEE, 2020, pp. 91--96.

\bibitem{liu2019fault}
T.~Liu, W.~Wen, L.~Jiang, Y.~Wang, C.~Yang, and G.~Quan, ``A fault-tolerant neural network architecture,'' in \emph{2019 56th ACM/IEEE Design Automation Conference (DAC)}.\hskip 1em plus 0.5em minus 0.4em\relax IEEE, 2019, pp. 1--6.

\bibitem{he2019noise}
Z.~He, J.~Lin, R.~Ewetz, J.-S. Yuan, and D.~Fan, ``Noise injection adaption: End-to-end reram crossbar non-ideal effect adaption for neural network mapping,'' in \emph{Proceedings of the 56th Annual Design Automation Conference 2019}, 2019, pp. 1--6.

\bibitem{yan2022swim}
Z.~Yan, X.~S. Hu, and Y.~Shi, ``Swim: Selective write-verify for computing-in-memory neural accelerators,'' in \emph{2022 59th ACM/IEEE Design Automation Conference (DAC)}.\hskip 1em plus 0.5em minus 0.4em\relax IEEE, 2022.

\bibitem{gao2021bayesian}
D.~Gao, Q.~Huang, G.~L. Zhang, X.~Yin, B.~Li, U.~Schlichtmann, and C.~Zhuo, ``Bayesian inference based robust computing on memristor crossbar,'' in \emph{2021 58th ACM/IEEE Design Automation Conference (DAC)}.\hskip 1em plus 0.5em minus 0.4em\relax IEEE, 2021, pp. 121--126.

\bibitem{shim2020two}
W.~Shim, J.-s. Seo, and S.~Yu, ``Two-step write--verify scheme and impact of the read noise in multilevel rram-based inference engine,'' \emph{Semiconductor Science and Technology}, vol.~35, no.~11, p. 115026, 2020.

\bibitem{yan2021uncertainty}
Z.~Yan, D.-C. Juan, X.~S. Hu, and Y.~Shi, ``Uncertainty modeling of emerging device based computing-in-memory neural accelerators with application to neural architecture search,'' in \emph{2021 26th Asia and South Pacific Design Automation Conference (ASP-DAC)}.\hskip 1em plus 0.5em minus 0.4em\relax IEEE, 2021, pp. 859--864.

\bibitem{yan2022radars}
Z.~Yan, W.~Jiang, X.~S. Hu, and Y.~Shi, ``Radars: Memory efficient reinforcement learning aided differentiable neural architecture search,'' in \emph{2022 27th Asia and South Pacific Design Automation Conference (ASP-DAC)}.\hskip 1em plus 0.5em minus 0.4em\relax IEEE, 2022, pp. 128--133.

\bibitem{jiang2020device}
W.~Jiang, Q.~Lou, Z.~Yan, L.~Yang, J.~Hu, X.~S. Hu, and Y.~Shi, ``Device-circuit-architecture co-exploration for computing-in-memory neural accelerators,'' \emph{IEEE Transactions on Computers}, vol.~70, no.~4, pp. 595--605, 2020.

\bibitem{yao2020fully}
P.~Yao, H.~Wu, B.~Gao, J.~Tang, Q.~Zhang, W.~Zhang, J.~J. Yang, and H.~Qian, ``Fully hardware-implemented memristor convolutional neural network,'' \emph{Nature}, vol. 577, no. 7792, pp. 641--646, 2020.

\bibitem{wang2022efficient}
Z.~Wang, C.~Huang, and Q.~Zhu, ``Efficient global robustness certification of neural networks via interleaving twin-network encoding,'' \emph{arXiv preprint arXiv:2203.14141}, 2022.

\bibitem{tsai2021formalizing}
Y.-L. Tsai, C.-Y. Hsu, C.-M. Yu, and P.-Y. Chen, ``Formalizing generalization and adversarial robustness of neural networks to weight perturbations,'' \emph{Advances in Neural Information Processing Systems}, vol.~34, 2021.

\bibitem{yan2022computing}
Z.~Yan, X.~S. Hu, and Y.~Shi, ``Computing in memory neural network accelerators for safety-critical systems: Can small device variations be disastrous?'' \emph{2022 International Conference on Computer-Aided Design (ICCAD)}, 2022.

\bibitem{yan2023improving}
Z.~Yan, Y.~Qin, X.~S. Hu, and Y.~Shi, ``Improving realistic worst-case performance of nvcim dnn accelerators through training with right-censored gaussian noise,'' \emph{2023 International Conference on Computer-Aided Design (ICCAD)}, 2023.

\bibitem{feinberg2018making}
B.~Feinberg, S.~Wang, and E.~Ipek, ``Making memristive neural network accelerators reliable,'' in \emph{2018 IEEE International Symposium on High Performance Computer Architecture (HPCA)}.\hskip 1em plus 0.5em minus 0.4em\relax IEEE, 2018, pp. 52--65.

\bibitem{wu2020adversarial}
D.~Wu, S.-T. Xia, and Y.~Wang, ``Adversarial weight perturbation helps robust generalization,'' \emph{Advances in Neural Information Processing Systems}, vol.~33, pp. 2958--2969, 2020.

\bibitem{chang2019nv}
C.-C. Chang, M.-H. Wu, J.-W. Lin, C.-H. Li, V.~Parmar, H.-Y. Lee, J.-H. Wei, S.-S. Sheu, M.~Suri, T.-S. Chang \emph{et~al.}, ``Nv-bnn: An accurate deep convolutional neural network based on binary stt-mram for adaptive ai edge,'' in \emph{2019 56th ACM/IEEE Design Automation Conference (DAC)}.\hskip 1em plus 0.5em minus 0.4em\relax IEEE, 2019, pp. 1--6.

\bibitem{chen2021pruning}
C.-Y. Chen and K.~Chakrabarty, ``Pruning of deep neural networks for fault-tolerant memristor-based accelerators,'' in \emph{2021 58th ACM/IEEE Design Automation Conference (DAC)}.\hskip 1em plus 0.5em minus 0.4em\relax IEEE, 2021, pp. 889--894.

\bibitem{shin2021fault}
H.~Shin, M.~Kang, and L.-S. Kim, ``Fault-free: A fault-resilient deep neural network accelerator based on realistic reram devices,'' in \emph{2021 58th ACM/IEEE Design Automation Conference (DAC)}.\hskip 1em plus 0.5em minus 0.4em\relax IEEE, 2021, pp. 1039--1044.

\bibitem{jeong2022variation}
S.~Jeong, J.~Kim, M.~Jeong, and Y.~Lee, ``Variation-tolerant and low r-ratio compute-in-memory reram macro with capacitive ternary mac operation,'' \emph{IEEE Transactions on Circuits and Systems I: Regular Papers}, 2022.

\bibitem{carlini2017towards}
N.~Carlini and D.~Wagner, ``Towards evaluating the robustness of neural networks,'' in \emph{2017 ieee symposium on security and privacy (sp)}.\hskip 1em plus 0.5em minus 0.4em\relax IEEE, 2017, pp. 39--57.

\bibitem{lecun1998gradient}
Y.~LeCun, L.~Bottou, Y.~Bengio, and P.~Haffner, ``Gradient-based learning applied to document recognition,'' \emph{Proceedings of the IEEE}, vol.~86, no.~11, pp. 2278--2324, 1998.

\bibitem{deng2012mnist}
L.~Deng, ``The mnist database of handwritten digit images for machine learning research,'' \emph{IEEE Signal Processing Magazine}, vol.~29, no.~6, pp. 141--142, 2012.

\bibitem{krizhevsky2009learning}
A.~Krizhevsky, G.~Hinton \emph{et~al.}, ``Learning multiple layers of features from tiny images,'' 2009.

\bibitem{he2016deep}
K.~He, X.~Zhang, S.~Ren, and J.~Sun, ``Deep residual learning for image recognition,'' in \emph{Proceedings of the IEEE conference on computer vision and pattern recognition}, 2016, pp. 770--778.

\bibitem{le2015tiny}
Y.~Le and X.~Yang, ``Tiny imagenet visual recognition challenge,'' \emph{CS 231N}, vol.~7, no.~7, p.~3, 2015.

\bibitem{deng2009imagenet}
J.~Deng, W.~Dong, R.~Socher, L.-J. Li, K.~Li, and L.~Fei-Fei, ``Imagenet: A large-scale hierarchical image database,'' in \emph{2009 IEEE conference on computer vision and pattern recognition}.\hskip 1em plus 0.5em minus 0.4em\relax Ieee, 2009, pp. 248--255.

\bibitem{simonyan2014very}
K.~Simonyan and A.~Zisserman, ``Very deep convolutional networks for large-scale image recognition,'' \emph{arXiv preprint arXiv:1409.1556}, 2014.

\end{thebibliography}

\begin{IEEEbiography}[{\includegraphics[width=1in,height=1.25in,clip,keepaspectratio]{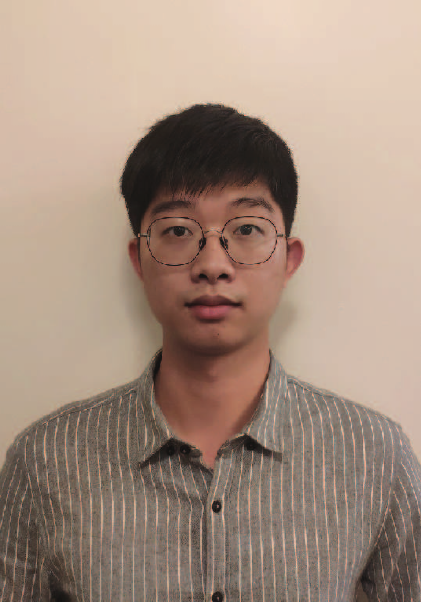}}]{Zheyu Yan} 
is a Ph.D. student in the Department of Computer Science and Engineering at the University of Notre Dame. He earned his B.S. degree from Zhejiang University in 2019. He is deeply interested in the co-design of software and hardware for deep neural network accelerators, particularly focusing on non-volatile memory-based compute-in-memory platforms.
\end{IEEEbiography}

\begin{IEEEbiography}[{\includegraphics[width=1in,height=1.25in,clip,keepaspectratio]{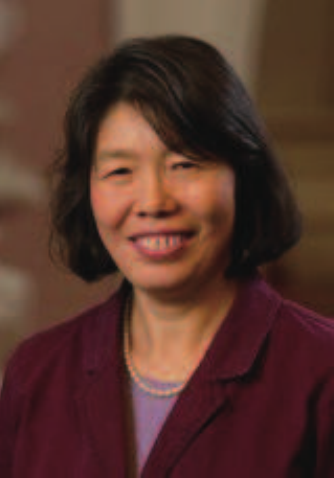}}]{Xiaobo Sharon Hu} 
earned her B.S. degree from Tianjin University in China in 1982, followed by an M.S. from the Polytechnic Institute of New York in 1984, and then a Ph.D. from Purdue University in West Lafayette, IN, USA, in 1989.

She currently holds a Professor position at the Department of Computer Science and Engineering at the University of Notre Dame, Notre Dame, IN, USA. Her research primarily revolves around beyond-CMOS technologies computing, low-power system design, and cyber-physical systems. Dr. Hu was honored with the NSF CAREER Award in 1997 and received Best Paper Awards from the Design Automation Conference in 2001 and the ACM/IEEE International Symposium on Low Power Electronics and Design in 2018.

In 2018, Dr. Hu took on the role of General Chair for the Design Automation Conference (DAC). She has been an Associate Editor for various publications including IEEE Transactions on Very Large Scale Integration (VLSI) Systems, ACM Transactions on Design Automation of Electronic Systems, and ACM Transactions on Embedded Computing. Currently, she serves as an Associate Editor for the ACM Transactions on Cyber-Physical Systems.
\end{IEEEbiography}

\begin{IEEEbiography}[{\includegraphics[width=1in,height=1.25in,clip,keepaspectratio]{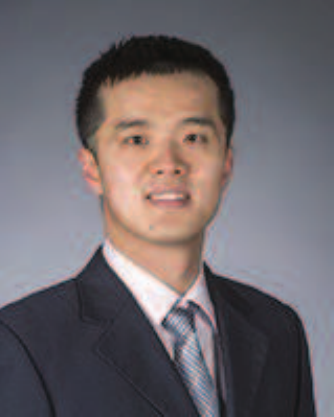}}]{Yiyu Shi} 
earned his B.S. degree with honors in electronic engineering from Tsinghua University in Beijing, China, in 2005. He later pursued his M.S. and Ph.D. degrees in electrical engineering at the University of California, Los Angeles, completing them in 2007 and 2009, respectively. He now serves as a Professor in the Departments of Computer Science and Engineering as well as Electrical Engineering at the University of Notre Dame, Notre Dame, IN, USA. His research primarily focuses on 3-D integrated circuits, hardware security, and applications in renewable energy. 

Prof. Shi has been recognized with multiple best paper nominations at premier conferences. He received the IBM Invention Achievement Award in 2009 and was honored with the Japan Society for the Promotion of Science Faculty Invitation Fellowship, the Humboldt Research Fellowship for Experienced Researchers, and the National Science Foundation CAREER Award. Additionally, he was the recipient of the IEEE Region 5 Outstanding Individual Achievement Award and the Air Force Summer Faculty Fellowship.
\end{IEEEbiography}

\end{document}